\documentclass[preprint,11pt,a4paper,authoryear]{elsarticle}
\usepackage{graphics,multirow,multicol}
\usepackage{geometry}
\usepackage[onehalfspacing]{setspace}
\usepackage{amssymb}
\usepackage{amsfonts}
\usepackage{amsmath}
\usepackage{amsthm}
\usepackage{natbib}
\usepackage{color,colortbl}
\usepackage[table,xcdraw,svgnames]{xcolor}
\usepackage[colorlinks]{hyperref}
\usepackage{graphicx}
\usepackage{epstopdf}
\usepackage{color, colortbl}
\usepackage{subfigure}
\usepackage{float}
\usepackage{bm,upgreek}
\usepackage[latin1]{inputenc}
\usepackage{rotating}
\usepackage{pdflscape}
\usepackage{textcomp}
\usepackage{eurosym}
\usepackage{xkeyval}
\usepackage{todonotes}
\usepackage{algorithm}
\usepackage{bbm}
\usepackage{pdflscape}
\usepackage{dsfont}
\usepackage[noend]{algpseudocode}
\usepackage{soul}
\usepackage{pgf}

\DeclareMathOperator*{\argmin}{arg\,min}
\newcommand{\ignore}[1]{}
\AtBeginDocument{
\hypersetup{
citecolor=blue,
linkcolor=blue,
urlcolor=Magenta}}

\numberwithin{equation}{section}

\newcommand\new[1]{}

\newtheorem{theorem}{Theorem}

\newtheorem{assumption}{Assumption}[section]

\renewcommand\appendixname{EC}
\let\emptyset\varnothing

\allowdisplaybreaks

\journal{\empty}
\begin{document}

\begin{frontmatter}

\title{Robust Classification via Support Vector Machines}

\author[IK]{Vali Asimit\corref{cor1}}
\ead{asimit@city.ac.uk}

\author[IK]{Ioannis Kyriakou}
\ead{ioannis.kyriakou@city.ac.uk}

\author[Simone]{Simone Santoni}
\ead{simone.santoni.1@city.ac.uk}

\author[SS]{Salvatore Scognamiglio}
\ead{salvatore.scognamiglio@uniparthenope.it}

\author[IK]{Rui Zhu}
\ead{rui.zhu@city.ac.uk}

\cortext[cor1]{Corresponding author: Vali Asimit (asimit@city.ac.uk)}

\address[IK]{Faculty of Actuarial Science \& Insurance, Bayes Business School, City, University of London, 106 Bunhill Row, London EC1Y 8TZ, UK}

\address[Simone]{Faculty of Management, Bayes Business School, City, University of London, 106 Bunhill Row, London EC1Y 8TZ, UK}

\address[SS]{Department of Business and Quantitative Studies, University of Naples ``Parthenope'', Generale Parisi Street, 80132 Naples, Italy}

\begin{abstract}
Classification models are very sensitive to data uncertainty, and finding robust classifiers that are less sensitive to data uncertainty has raised great interest in the machine learning literature.  This paper aims to construct robust \emph{Support Vector Machine} classifiers under feature data uncertainty via two probabilistic arguments. The first classifier, \emph{Single Perturbation}, reduces the local effect of data uncertainty with respect to one given feature and acts as a local test that could confirm or refute the presence of significant data uncertainty for that particular feature. The second classifier, \emph{Extreme Empirical Loss}, aims to reduce the aggregate effect of data uncertainty with respect to all features, which is possible via a trade-off between the number of prediction model violations and the size of these violations. Both methodologies are computationally efficient and our extensive numerical investigation highlights the advantages and possible limitations of the two robust classifiers on synthetic and real-life data.
\end{abstract}

\begin{keyword}
Analytics \sep support vector machine\sep robust classification\sep data uncertainty\sep extreme empirical loss.
\end{keyword}

\end{frontmatter}


\section{Introduction\label{intro sec}}

A prediction model is said to be robust if small changes in the data would
not change the model outputs, which is a consistent interpretation of robust
prediction modeling across the theoretical statistics and computational
robustness literature. Standard practical approaches to achieve robust
predictions in the machine learning literature include outlier detection,
testing prediction performance under data contamination or allowing for 
\emph{data uncertainty (DU)} in the prediction model. 

Creating robust prediction models is imperative whenever the current and/or
future data are affected by DU. Specifically, DU means that the data are
affected by i) sampling error, ii) ambiguity or iii) data noise, and each
source of DU affects the robustness of the prediction model. \emph{Sampling
error} is inevitable, and this source of DU is negligible in large samples,
which is often the case in machine learning modeling. \emph{Data ambiguity}
goes beyond missing data, and is a challenging issue that often occurs in
categorical data. For example, insurance fraud data contain information
about the event leading to a claim (major/minor) that is very subjective;
similarly, likert scale features in automatized hiring processes are
influenced by ambiguous questionnaires. \emph{Data noise} is the unexplained
variability within the observational data. The data noise could be
associated with the features or response variable(s) in supervised learning.
In summary, data ambiguity and feature and/or label noise are the important
sources of DU, though data noise is the main source of DU considered in the
robust classification literature.

\emph{Support vector machine (SVM)} is an effective classifier with a
variety of real-world applications. 
However, SVM is very sensitive to label and feature noise, and thus, a large amount of work has been done to robustify SVM against such sources of DU. For example, prior studies have
dealt with allocating different weights to individual data points to reduce
the effect of outliers~\citep{wu2013adaptively,yang2007weighted}; various
loss functions have been used to robustify SVM predictions with respect to feature
noise~\citep{bamakan2017ramp,brooks2011support,huang2013support,shen2017support,singh2014Closs,suykens1999least,wu2007robust}; standard approaches have also been explored in the field of \emph{robust optimization (RO)}, such as metric-type (non-probabilistic) uncertainty sets~\citep{Bertimas_et_al_2018,bi2005support,trafalis2006robust,xu2009robustness}; another RO approach, namely, \emph{chance constraints (CC)} that are probabilistic uncertainty sets, has been introduced in the literature so that the underlying optimization problems used in SVM prediction exhibit more robust outputs whenever DU is present~\citep{huang2012CC,lanckriet2002CC,Wang_et_al_2018}, though a wider discussion of SVM modeling with CC is available in \cite{Ben_Tal_et_al_2011}. It is evident -- even from this short introduction -- that \emph{Operational Research (OR)} has been a key ingredient in supervised and unsupervised learning, and an excellent review in this sense is available in \cite{Gambella_et_al_2021_EJOR}.

The main purpose of this paper is to reduce the effect of feature noise for
binary SVM classifiers. We explore the internal structure of the \emph{classical SVM (C-SVM)} classifier in order to detect and tackle the feature
noise via probabilistic arguments, while the eventual label noise issue is
put aside in this paper, since our proposed probabilistic arguments are not
easily extendable to prediction modeling under label noise.

The first proposed robust SVM classifier is the \emph{Single Perturbation
(SP-SVM)}, and aims to reduce the local effect of DU with respect to one
given feature by embedding this possible source of uncertainty into the
prediction model. This technique is very popular in RO, where the so-called
CCs are constructed to replace its non-robust counterparts (that are assumed
to be certain). SP-SVM is an effective classifier whenever DU is present,
and it could be used to test whether or not each feature is affected by DU. 

The second proposed robust SVM classifier is the \emph{Extreme Empirical
Loss (EEL-SVM)}, and aims to reduce the aggregate effect of DU. This is
achieved by introducing a trade-off between the number of prediction model
violations (misclassifications) and the size of these violations, which is
explained via a probabilistic argument. Simply speaking, C-SVM assigns the
same importance (probability) to each misclassification and aims to reduce
the overall prediction error. In contrast, EEL-SVM focuses only on the most
significant errors, i.e. the extreme errors, which increases the accuracy
around the borderline decisions and improves the model accuracy. This approach is inspired by the standard risk measure, namely, \emph{Conditional Value-at-Risk}, which was shown to be very robust in the context of model ambiguity modeled via RO \citep{asimit2017RO}.

C-SVM is a special case of both SP-SVM and EEL-SVM, and thus, our robust
formulations generalize C-SVM. Efficient convex quadratic programming
solvers are used for solving SP-SVM and EEL-SVM, but the computational times
of SP-SVM and EEL-SVM are not smaller than C-SVM, and this effect is known
in the RO literature as the \emph{price of robustness}. Note that, due to
its sparsity, EEL-SVM has a lower computational time than SP-SVM and the
reduction in computational time is mainly influenced by the sample size. Our
numerical experiments have shown that both SP-SVM and EEL-SVM perform very
well when compared with their robust SVM competitors. Moreover, even though
the overall performance of SP-SVM is slightly superior to EEL-SVM, the
sparsity trait of EEL-SVM is extremely appealing from the computational
perspective; for specific details about our numerical experiments, see
Section~\ref{num sec}.

It is worth noting that, unlike the existing methods based on CC~
\citep{Ben_Tal_et_al_2011,huang2012CC,lanckriet2002CC,Wang_et_al_2018}, SP-SVM does not require
estimating the covariance matrix, which is not always computationally stable~
\citep{fan2016overview,ledoit2020power}. Moreover, in this paper, we only provide
explicit derivations for SP-SVM and EEL-SVM with the most popular loss
function, i.e. Hinge loss, but similar derivations are possible for any
other loss function. 
In addition to the two new robust SVM models, we also prove Fisher
consistency for a general convex loss function and binary SVM classifier.

The paper is organized as follows. Section~\ref{backg pb def sec} provides
the necessary background, while Section~\ref{robust svm sec} illustrates the
two proposed robust SVM classifiers. Section~\ref{num sec} summarizes our
numerical experiments conducted over synthetic and real-life datasets. 
Section~\ref{concl sec} includes some final comments and recommendations that emerge from our paper.  All proofs are relegated to the appendix.

\section{Background and Problem Definition\label{backg pb def sec}}

The current section takes stock of the necessary background related to
binary SVM classification. Section~\ref{subsec:pb_def} briefly explains the
C-SVM formulation, while Section~\ref{subsec:loss functions} provides a
comprehensive description of the pros and cons of various loss functions
which is a pivotal choice for any SVM implementation. Finally, Section~\ref%
{subsec:Fisher} discusses the importance of Fisher consistency property in
classification followed by a theoretical contribution in the context of
binary classification, which is stated as Theorem~\ref{Fisher class SVM th}.

\subsection{Problem Definition\label{subsec:pb_def}}

Our starting point is the training set that contains $N$ instances and their
associated labels, $\{(\mathbf{x}_{i},y_{i}),\;i=1,\ldots ,N\}$, where $%
\mathbf{x}_{i}\!\in \!\mathcal{X}\!\subseteq \!\mathbb{R}^{d}$ and $%
y_{i}\!\in \!\mathcal{Y}$. The training set is assumed to be sampled from $%
(X,Y)$, but the binary classification reduces to $\mathcal{Y}:=\{-1,1\}$,
where $y_{i}=1$ if $\mathbf{x}_{i}$ is in the positive class, $\mathcal{C}%
_{+1}$, and $y_{i}=-1$ if $\mathbf{x}_{i}$ is in the negative class, $%
\mathcal{C}_{-1}$. The main objective is to construct an accurate (binary)
classifier $c:\mathcal{X}\rightarrow \mathcal{Y}$ which maximizes the
probability that $c\big(\mathbf{x}_{i}\big)=y_{i}$.

SVM aims to identify a separation hyper-plane $\mathbf{w}^{T}\phi \big(%
\mathbf{x}\big)+b$ that generates two parallel supporting hyper-planes:%
\begin{equation}
\mathbf{w}^{T}\phi \big(\mathbf{x}\big)+b=1\quad \mbox{and}\quad \mathbf{w}
^{T}\phi \big(\mathbf{x}\big)+b=-1,  \label{SVM hyper-planes}
\end{equation}%
where $\phi (\cdot )$ is a notional function that transforms the feature
space into a synthetic feature space that allows a linear hyper-plane
separation of the data (when linear classifiers are not effective on the
original data). The data are rarely perfectly separable, and a compromise is
made by allowing classification violations for the non-separable data. The
latter is also known as \emph{soft-margin SVM} and is formulated as follows: 
\begin{equation}
\underset{\mathbf{w},b}{\text{min}}\;\frac{1}{2}\mathbf{w}^{T}\mathbf{w}+C %
\displaystyle\sum_{i=1}^{N}L(1-y_{i}(\mathbf{w}^{T}\phi (\mathbf{x}_{i})+b)).
\label{SVM opt pb}
\end{equation}
The first term aims to find the `best' classifier by maximizing the distance
between the two hyper-planes defined in \eqref{SVM hyper-planes}, while the
second term penalizes the classifier's violations measured via a given loss
function $L:\mathbb{R} \rightarrow \mathbb{R} _{+}$; for details, see \cite{
vapnik2}.

\subsection{Loss Function\label{subsec:loss functions}}

Solving the general SVM formulation in \eqref{SVM opt pb} requires specific
solvers that depend upon the loss function choice, which has a central role
in SVM classification. The existing literature has dealt with numerous
piecewise loss functions, and a summary is given below:

\begin{enumerate}
\item[i)] \textit{Hinge loss:} $L_{H}(u):=\max\{0,u\}$; 

\item[ii)] \textit{Truncated Hinge loss ($a\ge 1$):} $L_{TH}(u):=\min\{\max%
\{0,u\},a\}$; 

\item[iii)] \textit{Pinball loss ($a\le0$):} $L_{P}(u):=\max\{au,u\}$; 

\item[iv)] \textit{Pinball loss with `$\epsilon$ zone' ($\epsilon\ge0$ and $%
a,b\le0$):} $L_{PEZ}(u):=\max\{0,u-\epsilon, a u+b\}$; 

\item[v)] \textit{Truncated Pinball loss ($a\le0$ and $b\ge0$):} $%
L_{TP}(u):=\max\{u,\min\{a u,b\}\}$.
\end{enumerate}

The standard choice, $L_{H}$, leads to efficient computations as it reduces %
\eqref{SVM opt pb} to solving a convex \textit{Linearly Constrained
Quadratic Program (LCQP)}, which is the original SVM formulation as
explained in \cite{vapnik2}. Moreover, the Hinge loss is proved to be an
upper bound of the classification error %
\citep{zhang2004statistical,shen2017support}, therefore it is a pivotal loss
choice. 
At the same time, the Hinge loss has been criticized for not being robust
and extremely sensitive to outliers, whereas the Truncated Hinge loss
proposed by \cite{wu2007robust} overcomes this issue at the expense of
computational complexity. The lack of convexity of this loss function
requires a bespoke algorithm, namely, the \textit{Difference of Convex
Functions Optimization Algorithm (DCA)}, which is computationally less
efficient than standard LCQP solvers used for solving C-SVM and our proposed
robust formulations (SP-SVM and EEL-SVM). The optimization problems based on
the two Pinball losses require solving LCQPs with many more linear equality
constraints than the Hinge loss, but the Pinball loss seems to be more
robust and stable when re-sampling~\citep{huang2013support}. Similar
arguments have been used in \cite{shen2017support} to justify that the
Truncated Pinball loss is a good choice when dealing with feature noise,
though it shares the same computational shortcoming with the Truncated Hinge
loss, that is, it requires non-convex solvers.

All previous five loss functions are piecewise linear, which is a
computational advantage, but non-piecewise linear loss functions have been
proposed in the existing literature. For example, the \emph{Least Square
loss} with $L_{LS}(u):=u^{2}$ is considered in \cite{suykens1999least} for
which an efficient LCQP formulation is proposed; the \emph{Correntropy loss} is defined in \cite{singh2014Closs}, but variants of this have been
investigated (see 
\citealp{xu2017robust}
for SVM-like formulations).
One could understand the possible advantages of non-linear convex loss
functions for other classification methods from \cite{lin2004note}, where
the Hinge loss is shown to be the tightest margin-based upper bound of the
misclassification loss for many well-known classifiers. Further, it is
numerically shown that this property does not suffice to think of the Hinge
loss as the universally `best' choice to measure misclassification. Strictly
convex loss functions are argued in \cite{bartlett2006convexity} to possess
appealing statistical properties when studying misclassification.

Even though the loss function is a pivotal choice in SVM classification, our
two new robust classifiers are not restricted to a specific loss function,
and thus, SP-SVM and EEL-SVM formulations are quite general and introduce
two new methodologies of tackling binary classification in the presence of
DU. However, the SP-SVM (instance \eqref{SVM non_sep DU}) and EEL-SVM
(instance \eqref{EEL-SVM def}) illustrations of this paper are only focused
on Hinge loss formulations, though any other illustrations are possible.
Therefore, the robustness of SP-SVM and EEL-SVM is a result of how these
classifiers deal with DU, and not a by-product of choosing the `best' loss
function.

\subsection{Fisher Consistency\label{subsec:Fisher}}

A desirable loss function property for a generic classifier is \emph{Fisher
consistency} or \emph{classification calibration}~%
\citep{bartlett2006convexity}. By definition, the loss function $L$ is
Fisher consistent if 
\begin{equation}
\argmin_{f:\mathcal{X}\rightarrow \mathbb{R}}\mathbf{E}_{\mathcal{X},%
\mathcal{Y}}L\big(1-Yf(\mathbf{X})\big)  \label{Fisher class def}
\end{equation}%
is solved by the Bayes classifier that is defined as follows: 
\begin{equation*}
f_{Bayes}^{\ast }(\mathbf{x})=\left\{ 
\begin{array}{rll}
1, & \quad \text{if}\; & \Pr (Y=1|\mathbf{x})>\Pr (Y=-1|\mathbf{x}), \\ 
-1, & \quad \text{if}\; & \Pr (Y=1|\mathbf{x})<\Pr (Y=-1|\mathbf{x}).%
\end{array}%
\right. 
\end{equation*}%
In the context of binary SVM classification, one could show that %
\eqref{Fisher class def} holds if 
\begin{eqnarray}\label{Fisher class SVM def} 
	\argmin_{z\in \mathbb{R} }\mathbf{E}_{\mathcal{Y}|\textbf{x}}L(1-Yz)=f_{Bayes}^*(\mathbf{x})
\end{eqnarray}
is true for all $\mathbf{x}\in \mathcal{X}$; e.g., see Proposition~1 in \cite%
{wu2007robust}.

Fisher consistency has been extensively investigated in the literature, and
we now provide a concise review that relates to our framework. Theorem~3.1
in \cite{lin2004note} shows that, if the global minimizer of \eqref{Fisher
class def} exists, then it has to be the same as the Bayes decision rule,
which is valid for any classification method. In the binary SVM setting,
Proposition~1 in \cite{wu2007robust} and Theorem~1 in \cite{shen2017support}
show that this property also holds for non-convex loss functions; the first
result covers a large set of truncated loss functions, whereas the second
focuses on the Truncated Pinball loss. Lemma~3.1 of \cite{lin2002support}
shows that the Hinge loss is Fisher consistent, and Theorem~1 in \cite%
{huang2013support} shows the same for the (convex loss) Pinball loss. Our
next result extends Fisher consistency to a general convex loss function $L$
for the binary SVM case, and its proof is relegated to Appendix~\ref%
{proof_th1}.

\begin{theorem}
\label{Fisher class SVM th} Assume that $L:\mathbb{R}\to\mathbb{R}_+$ is a
convex loss function such that $L(0)=0$. If $L(\cdot)$ is linear on $%
(0,2+\epsilon)$ for some $\epsilon>0$, then $L$ is Fisher consistent.
\end{theorem}

\section{Robust SVM\label{robust svm sec}}

This section explains the concept of robust SVM in a more formal way and
provides the technical 
details for our two robust SVM; that is, SP-SVM in Section~\ref{robust svm
sec: def}, and EEL-SVM in Section~\ref{robust svm sec: eel}. Finally, a
summary of practical recommendations when using our robust SVM formulations
is given in Section~\ref{robust svm sec: recom}.



\subsection{Single Perturbation SVM\label{robust svm sec: def}}

SP-SVM aims to reduce the local effect of DU with respect to one given
feature by embedding this possible source of uncertainty into the
optimization problem that describes our robust prediction model. The main
idea is the appropriation of a standard RO approach that relies on CCs
constructed to replace its non-robust counterparts (assumed to hold almost
surely). That is, SP-SVM extends C-SVM by adjusting the feasibility set
through parameter $\alpha $ that controls the DU level; $\alpha $ is tuned
like in any other hyper-parameter model.

Previous robust SVM classifiers that rely on CC require the covariance
(feature) matrix estimate, which is computable but brings some practical
drawbacks; the empirical covariance matrix is computationally unstable when $%
d$ is large, and is often not positive semi-definite if missing values are
present in the sample. This is not the case for SP-SVM, whose robustness is
achieved by identifying the feature affected the most by DU. We choose this
feature as the one with the highest variance whenever data are less
interpretable (see Sections~\ref{num sec: sim data} and \ref{num sec: real
data}), though domain knowledge could help with choosing this feature (see
Section~\ref{fin_data_sec}).

Solving \eqref{SVM opt pb} under the Hinge loss is equivalent to solving the
following LCQP instance:%
\begin{eqnarray}
&&\underset{\mathbf{w},b,\boldsymbol{\xi }}{\text{min}}\;\;\frac{1}{2}%
\mathbf{w}^{T}\mathbf{w}\!+\!C\displaystyle\sum_{i=1}^{N}\xi _{i}
\label{SVM non_sep classical} \\
&&\;\text{s.t.}\;\;y_{i}\big(\mathbf{w}^{T}\phi (\mathbf{x}_{i})\!+\!b\big)%
\geq 1\!-\!\xi _{i},\;\xi _{i}\geq 0,\;\;1\leq i\leq N,  \notag
\end{eqnarray}%
where $C>0$ is a penalty constant that becomes a tuning parameter in the
actual implementation phase. Equation~\eqref{SVM non_sep classical}
represents the mathematical formulation of C-SVM.

We are interested in calibrating \eqref{SVM non_sep classical} in the
presence of DU with respect to one feature, e.g., the $k^{th}$ feature.
Thus, the $j^{th}$ entry of $\phi \big(\mathbf{x}_{i}\big)$, denoted by $%
\phi _{j}\big(\mathbf{x}_{i}\big)$, is deterministic for all $1\leq i\leq N$
and $1\leq j\neq k\leq d$, whereas the $k^{th}$ feature is affected by an
error term $Z_{ik}$, hence $\phi _{k}\big(\mathbf{x}_{i}\big)$ is replaced
by $\phi _{k}\big(\mathbf{x}_{i}\big)+Z_{ik}$ for all $1\leq i\leq N$.
Moreover, each error term is defined on a probability space $(\Omega _{ik},%
\mathcal{F},P)$ with $\Omega _{ik}\subseteq \mathbb{R}$. Therefore, the DU
variant of \eqref{SVM non_sep classical} with respect to the $k^{th}$
feature becomes 
\begin{eqnarray}
&&\underset{\mathbf{w},b,\boldsymbol{\xi }}{\text{min}}\;\;\frac{1}{2}%
\mathbf{w}^{T}\mathbf{w}+C\displaystyle\sum_{i=1}^{N}\xi _{i}
\label{SVM non_sep DU} \\
&&\;\text{s.t.}\;\;\Pr \bigg(y_{i}\Big(\mathbf{w}^{T}\phi \big(\mathbf{x}_{i}%
\big)+w_{k}Z_{ik}+b\Big)\geq 1-\xi _{i}\bigg)\geq \alpha ,\;\xi _{i}\geq
0,\;1\leq i\leq N,  \notag
\end{eqnarray}%
where $\alpha \in \lbrack 0,1]$ reflects the unknown modeler's perception of
DU that is later tuned. This kind of probability-like constraint is also
known as CC in the OR literature.

For any given tuple $(i,k)$, the \textit{cumulative distribution function
(cdf)} of $Z_{ik}$, $F_{ik}(\cdot )$, is defined on $\Omega _{ik}$.
Furthermore, we define two generalized inverse functions as follows: 
\begin{equation*}
F_{ik}^{-1}(t):=\inf \big\{x\in \mathbb{R}:\;F_{ik}(x)\geq t\big\}\quad 
\text{and}\quad F_{ik}^{-1+}(t):=\sup \big\{x\in \mathbb{R}:\;F_{ik}(x)\leq t%
\big\}
\end{equation*}%
for all $t\in \lbrack 0,1]$, where $\inf \emptyset =\infty $ and $\sup
\emptyset =-\infty $ hold by convention. Clearly, 
\begin{equation*}
t\leq \Pr \big(Z_{ik}\leq x\big)\Leftrightarrow F_{ik}^{-1}(t)\leq x\quad 
\text{and}\quad \Pr \big(Z_{ik}<\!x\big)\!\leq t\Leftrightarrow x\leq
F_{ik}^{-1+}(t),\;\;x\in \mathbb{R}\;\text{and}\;t\in \lbrack 0,1].
\end{equation*}%
Therefore, the CC from \eqref{SVM non_sep DU} is equivalent to 
\begin{equation}
y_{i}\big(\mathbf{w}^{T}\phi (\mathbf{x}_{i})+b\big)%
+y_{i}w_{k}F_{ik}^{-1+}(1\!-\!\alpha )\geq 1\!-\!\xi _{i},\;\;y_{i}\big(%
\mathbf{w}^{T}\phi (\mathbf{x}_{i})+b\big)+y_{i}w_{k}F_{ik}^{-1}(\alpha
)\geq 1\!-\!\xi _{i}  \label{CC: eq1}
\end{equation}%
when $y_{i}w_{k}\geq 0$ or $y_{i}w_{k}<0$, respectively. Without imposing
any restriction on $F_{ik}$, the conditional constraint from \eqref{CC: eq1}
makes \eqref{SVM non_sep DU} a mixed integer programming instance, which is
a major computational shortcoming for large scale problems. The next set of
conditions enable us to solve \eqref{SVM non_sep DU} efficiently.

\begin{assumption}
\label{ass1} $F_{ik}^{-1}(\alpha)+F_{ik}^{-1+}(1-\alpha )=0$ for a given
integer $1\leq k\leq d$ and some $\alpha \in \lbrack 0,1]$.
\end{assumption}

If the random error $Z_{ik}$ is defined on $\Omega _{ik}=\big(-\omega
_{ik},\omega _{ik}\big)$ with $0<\omega _{ik}\leq \infty $ such that its cdf
is continuous and increasing, and $F_{ik}(\cdot )+F_{ik}(-\cdot )=1$ in a
neighborhood of $F_{ik}^{-1+}(\alpha )$, then Assumption~\ref{ass1} holds.
Note that symmetric and continuous cdf, such as Gaussian, Student's $t$ or
any other member of the elliptical family of distributions centered at $0$
satisfies Assumption~\ref{ass1}; for details, see~\cite{fang1990}.
Therefore, Assumption~\ref{ass1} is quite general.

Under Assumption~\ref{ass1}, \eqref{CC: eq1} is equivalent to 
\begin{equation*}
y_{i}\big(\mathbf{w}^{T}\phi (\mathbf{x}_{i})+b\big)-|w_{k}|F_{ik}^{-1}(%
\alpha )\geq 1-\xi _{i},
\end{equation*}%
and in turn, \eqref{SVM non_sep DU} is equivalent to solving 
\begin{equation}
\begin{array}{llc}
\underset{\mathbf{w},b,\boldsymbol{\xi }}{\text{min}} & \frac{1}{2}\mathbf{w}%
^{T}\mathbf{w}+C\displaystyle\sum_{i=1}^{N}\xi _{i} &  \\ 
\;\text{s.t.} & y_{i}\big(\mathbf{w}^{T}\phi (\mathbf{x}_{i})+b\big)\geq
1\!-\!\xi _{i} & (i) \\ 
& y_{i}\big(\mathbf{w}^{T}\phi (\mathbf{x}_{i})+b\big)-y_{i}w_{k}a_{ik}\geq
1\!-\!\xi _{i} & (ii) \\ 
& y_{i}\big(\mathbf{w}^{T}\phi (\mathbf{x}_{i})+b\big)+y_{i}w_{k}a_{ik}\geq
1\!-\!\xi _{i} & (iii) \\ 
& \xi _{i}\geq 0,\;\;1\leq i\leq N, & (iv)%
\end{array}
\label{SVM non_sep simple}
\end{equation}%
where $a_{ik}:=F_{ik}^{-1}(\alpha )$. If $a_{ik}\leq 0$ for all $1\leq i\leq
N$, then the inequality constraints \eqref{SVM non_sep simple}~(ii) and
(iii) are redundant, and thus, SP-SVM and SVM are identical, i.e. the $k^{th}
$ feature is not affected by DU. If $a_{ik}\geq 0$ for all $1\leq i\leq N$,
then the inequality constraint \eqref{SVM non_sep simple}~(i) is redundant
and the $k^{th}$ feature is affected by DU, in which case SP-SVM becomes
more conservative than SVM, i.e. the SP-SVM hyper-plane violations $\xi _{i}$%
's are allowed to be larger (than the C-SVM violations) due to DU.

We now provide a practical recommendation to finding a `reasonable' choice
for $a_{ik}$. One possibility is to assume a Gaussian random noise with zero
mean and variance equal to the sampling error estimate, i.e. 
\begin{equation*}
a_{ik}=\hat{a}_{k}:=q_{\alpha ,G}\sqrt{\frac{1}{N\!-\!1}\displaystyle%
\sum_{i=1}^{N}(x_{ik}-\bar{x}_{k})^{2}}\;\mbox{and}\;\bar{x}_{k}:=\frac{1}{N}%
\sum_{i=1}^{N}x_{ik}\quad \text{for all $1\leq i\leq N$},
\end{equation*}%
where $q_{\alpha ,G}$ is the $\alpha $-Normal quantile. It could be argued
that the Gaussian random noise might underestimate DU, hence a more
heavy-tailed noise, such as Student's $t$, might be more appropriate; in
that case, we could simply replace $q_{\alpha ,G}$ by the $\alpha $-quantile
of the distribution of choice. We always tune $\alpha $ for values greater
than $0.5$, i.e. $a_{ik}>0$ for all $1\leq i\leq N$, since DU is assumed to
be present. Therefore, our SP-SVM implementations require solving LCQP
instances with $3N$ inequality constraints, though C-SVM implementations
require solving LCQP instances with $2N$ inequality constraints; this is not
surprising and is known as the price of robustness in the OR literature. An explicit solution for \eqref{SVM non_sep simple} is
detailed in Appendix~\ref{sol_SP_SVM}, where we do not make any assumption
on the sign of $a_{ik}$'s.

\subsection{Extreme Empirical Loss SVM\label{robust svm sec: eel}}

EEL-SVM is designed to reduce the overall effect of DU with respect to all
features that are possibly affected by random noise, which is different from
SP-SVM where only one feature is assumed to be affected by noise. This does
not mean that EEL-SVM is `better' than SP-SVM, as the two approaches
complement each other, and Section~\ref{num sec} provides empirical evidence
in that sense.

Simply speaking, EEL-SVM creates a trade-off between the number of model
misclassifications and the size of these violations, which is explained via
a probabilistic argument. While C-SVM assigns the same importance
(probability) to each $\xi_i$, i.e. $1/N$, so that the overall prediction
error is minimized, EEL-SVM considers that only some of the largest
individual model violations affect the overall classification error. This
means that EEL-SVM robustifies the classifier by paying particular attention
to outliers without removing such data points that go astray from the
general trend, since such sub-samples may not be a negligible portion of the
data when DU is present.

The soft-margin SVM from \eqref{SVM opt pb} could be rewritten as 
\begin{equation}
\underset{\mathbf{w},b}{\text{min}}\;\frac{1}{2}\mathbf{w}^{T}\mathbf{w}+C%
\widehat{E}\Big[L\Big(1-Y\big(\mathbf{w}^{T}\phi \big(\mathbf{X}\big)+b\big)%
\Big)\Big],  \label{SVM opt pb error}
\end{equation}%
where the second term acts as the empirical estimate of the penalty
associated with the classifier's violations; this is given by the average
model deviation measured via the loss function $L$. The loss function choice
could influence the borderline decisions where examples could be classified
either way, and thus, a `good' loss choice may reduce the misclassification
error. Many SVM classifiers focus on the choice of $L$, though the penalty
term is always based on the usual sample average with equal importance to
all hyper-plane violations. EEL-SVM aims to focus more on the large
deviations that may considerably perturb the classification decision in the
presence of DU. To this end, we place more weight to the larger violations
via a novel empirical penalty function, namely, the \textit{Extreme
Empirical Loss (EEL)}, which is formulated as 
\begin{equation}
\min_{z}z\!+\!\frac{1}{N(1\!-\!\alpha )}\sum_{i=1}^{N}\max \big\{\zeta
_{i}\!-\!z,0\big\},  \label{EEL def}
\end{equation}%
where $\zeta _{i}=L\big(1-y_{i}(\mathbf{w}^{T}\phi (\mathbf{x}_{i})\!+\!b)%
\big)$ are the individual model violations. Note that \eqref{EEL def} is the
empirical estimate of the \textit{Conditional Value-at-Risk at level $\alpha 
$ ($CVaR_{\alpha }$)} of the classifier's violations, i.e. 
\begin{equation*}
\widehat{CVaR}_{\alpha }\Big(L\big(1-Y(\mathbf{w}^{T}\phi (\mathbf{X})+b)%
\big)\Big).
\end{equation*}%
For details, see the seminal paper \cite{cvar200O} which introduces $CVaR$,
a well-known risk management measure. The parameter $0\leq \alpha <1$
represents the caution level chosen by the modeler; a higher value of $%
\alpha $ would penalize fewer extreme violations, i.e. large $\xi _{i}$'s. 
This is made obvious by noting that \eqref{EEL def} is equivalent to 
\begin{equation*}
\frac{1}{r}\sum_{i=1}^{r}\zeta _{i,N}\;\mbox{if}\;\alpha =1-\frac{r}{N}%
,\;1\leq r\leq N
\end{equation*}%
for any integer $r$, where $\zeta _{1,N}\geq \zeta _{2,N}\geq \cdots \geq
\zeta _{N,N}$ are the upper order statistics of the sample $\big\{\zeta
_{i};1\leq i\leq N\big\}$. Clearly, the least conservative EEL is attained
when $\alpha =0$, and becomes the sample average $\frac{1}{N}%
\sum_{i=1}^{N}\zeta _{i}$, meaning that C-SVM is a special case of EEL-SVM
(when $\alpha =0$).


Keeping in mind \eqref{SVM opt pb} and \eqref{EEL def}, the
mathematical formulation of EEL-SVM is equivalent to solving the instance%
\begin{equation*}
\begin{array}{ll}
\underset{\mathbf{w},b,z,\boldsymbol{\xi }}{\text{min}} & \frac{1}{2}\mathbf{%
w}^{T}\mathbf{w}+Dz\!+\!\displaystyle\frac{D}{N(1\!-\!\alpha )}%
\sum_{i=1}^{N}\xi _{i} \\ 
\;\;\;\text{s.t.} & \xi _{i}+z\geq L\big(1-y_{i}(\mathbf{w}^{T}\phi (\mathbf{%
x}_{i})\!+\!b)\big),\;\xi _{i}\geq 0,\;\;1\leq i\leq N,%
\end{array}%
\end{equation*}%
for any loss function $L$, while the Hinge loss choice simplifies EEL-SVM to
solving the convex LCQP instance from below%
\begin{equation}
\begin{array}{ll}
\underset{\mathbf{w},b,z,\boldsymbol{\xi }}{\text{min}} & \frac{1}{2}\mathbf{%
w}^{T}\mathbf{w}+Dz\!+\!\displaystyle\frac{D}{N(1\!-\!\alpha )}%
\sum_{i=1}^{N}\xi _{i} \\ 
\;\;\;\text{s.t.} & y_{i}(\mathbf{w}^{T}\phi (\mathbf{x}_{i})\!+\!b)\!+\!z%
\geq 1-\xi _{i},\;\xi _{i}+z\geq 0,\;\xi _{i}\geq 0,\;\;1\leq i\leq N.%
\end{array}
\label{EEL-SVM def}
\end{equation}%
Here, $D>0$ is a penalty constant that becomes a tuning parameter in the
actual implementation, which has a similar purpose as the penalty constant $C
$ in \eqref{SVM non_sep simple}. One may derive similar formulations for any
other loss function and easily write the convex LCQP formulations for
Pinball loss and Pinball loss with `$\epsilon $ zone'; non-convex loss
functions, such as Truncated Hinge and Truncated Pinball, require bespoke
DCA solvers, but such details are beyond the scope of this paper. The
explicit solution of \eqref{EEL-SVM def} is given in Appendix~{\ref%
{sol_EEL_SVM}} via the usual duality arguments. Note that the convex
instance \eqref{EEL-SVM def} requires solving LCQP with $3N$ inequality
constraints, which has the same computational complexity as SP-SVM, though
EEL-SVM is more sparse.

\subsection{Recommendations Related to the Use of the Two New Formulations
\label{robust svm sec: recom}}

We first summarise the traits of SP-SVM and EEL-SVM. SP-SVM has a local
robust treatment and focuses on the mostly affected feature by DU,
identified in Section~\ref{num sec} via variance, though domain knowledge
could be useful in determining that feature. EEL-SVM does not differentiate
among features and proposes an overall robust treatment by finding a
trade-off between the number of significant (to the prediction model)
extreme violations and the level of these violations. 

Let us anticipate the computational pros and cons of SP-SVM and EEL-SVM. Our
numerical experiments in the next section show that the overall performance
of SP-SVM and EEL-SVM are comparable. Both generalize C-SVM at the expense
of computational cost, known as the price of robustness. Moreover, the
computational time for EEL-SVM is marginally lower than for SP-SVM due to
the sparsity of the former. 

\section{Numerical Experiments\label{num sec}}

In this section, we conduct all our numerical experiments that compare our
robust classifiers (SP-SVM and EEL-SVM) with four other SVM classifiers by
checking the classification accuracy and robustness resilience. The four SVM
competitors include C-SVM~{\citep{vapnik1}} and three well-known robust SVM
classifiers, i.e.

\begin{itemize}
\item[i)] \emph{Pinball SVM ($pin$-SVM)} -- see \cite{huang2013support}; 

\item[ii)] \emph{Truncated Pinball SVM ($\overline{pin}$-SVM)} -- \cite%
{shen2017support}; 

\item[iii)] \emph{Ramp Loss K-Support Vector Classification-Regression
(Ramp-KSVCR)} -- see \cite{bamakan2017ramp}.
\end{itemize}

Note that the three classifiers above build up robust decision rules by
modifying the standard Hinge loss used in C-SVM. Classifiers i) and ii) are
based on their corresponding loss functions listed in Section~\ref%
{subsec:loss functions}, whereas classifier iii) relies on a mixture of loss
functions.

Section~\ref{num sec: sim data} consists of a data analysis based on
synthetic data, where the `true' classification decision has a closed-form.
Section~\ref{num sec: real data} compares all six binary classifiers over
various widely investigated real-life datasets. Section~\ref{fin_data_sec}
offers a more qualitative analysis of SVM robust classification, where it is
explained how DU can be identified so that one can validate whether a robust
classifier is fit for purpose in practice. The code could be retrieved from a public repository that is available at
\url{https://github.com/salvatorescognamiglio/SPsvm_EELsvm}.

\subsection{Synthetic Data\label{num sec: sim data}}

The first set of numerical experiments compares the classification
performance of SP-SVM, EEL-SVM, C-SVM, $pin$-SVM and $\overline{pin}$-SVM
for simulated data generated as in \cite{huang2013support} and \cite%
{shen2017support}. Note that we are not able to compare these five SVM
classifiers with Ramp-KSVCR, since the publicly available code for it does
not report the tuned separation hyper-plane parameters, though the
classification performances of all six classifiers (including Ramp-KSVCR)
are compared in Section~\ref{num sec: real data} in terms of accuracy and
robustness resilience to contamination.

We do not assume DU in Section~\ref{num sec: sim data - no noise}, and data
contamination is added in Section~\ref{num sec: sim data - with noise}. The
non-contaminated data are simulated based on a Gaussian bivariate model for
which the analytical/theoretical or `true' linear classification boundary is
known (referred to as \emph{Bayes classifier} from now on). Further, nested
simulation is used to generate the labels via a Bernoulli random variable $B$
with probability of `success' $p=0.5$; therefore, we generate $N\in
\{100,200\}$ random variates from this distribution, where $N$ is the total
number of examples from the two classes. The features $\{\mathbf{x}%
_{i},\;i=1,\ldots ,N\}$ are simulated according to 
\begin{equation}
\mathbf{X}_{i}\!\mid B\!=\!1\sim \mathcal{N}(\boldsymbol{\mu },\boldsymbol{%
\Sigma }),\;\mathbf{X}_{i}\!\mid \!B\!=\!-1\sim \mathcal{N}(\!-\boldsymbol{%
\mu },\boldsymbol{\Sigma }),\;\text{$\boldsymbol{\mu }=[0.5,-3]^{T}$ and $%
\boldsymbol{\Sigma }=\mathrm{diag}(0.2,3)$}.  \label{sampling}
\end{equation}%
Note that we estimate in Sections~\ref{num sec: sim data - no noise} and \ref{num sec: sim data - with noise} the classifiers for these synthetic data, i.e. $x_{2}=m x_{1}+q$, and compare with the Bayes classifier, i.e. $x_{2}=m_{0}x_{1}+q_{0}$ where $m_{0}=2.5$ and $q_{0}=0$. SP-SVM training is performed by considering DU
only with respect to the second feature that has a higher variance. 

\subsubsection{Synthetic Non-contaminated Data\label{num sec: sim data - no
noise}}

The data are simulated and we conduct a $10$-fold cross-validation to tune
the parameters of each classifier over the following parameter spaces:

\begin{itemize}
\item SP-SVM: $\alpha \in \mathcal{A}_{SP}=\{0.50,0.51,\dots,0.60\}$; 

\item EEL-SVM: $\alpha \in \mathcal{A}_{EEL}=\{0,0.01,0.02\}$; 

\item $pin$-SVM: $\tau \in \mathcal{T}_{pin}=\{0.1,0.2,\dots,1\}$; 

\item $\overline{pin}$-SVM: $(\tau,s)\in \mathcal{T}_{\overline{pin}}\times 
\mathcal{S}$, where $\mathcal{T}_{\overline{pin}}=\{0.25,0.5,0.75\}$ and $%
\mathcal{S}:=\{0.25,0.5,0.75, 1\}$.
\end{itemize}

\noindent In the interest of fair comparisons, the parameter spaces have similar
cardinality, except EEL-SVM that has a smaller-sized parameter space, which
does not create any advantage to EEL-SVM. We choose the same penalty value
for all methods by setting $C=100$ (for C-SVM, SP-SVM, $pin$-SVM and $%
\overline{pin}$-SVM) and $D=100\times N$ (for EEL-SVM).

The five SVM classifiers are compared via $100$ independent samples of size $%
N$ for which $(m_{i},q_{i})$ is computed for all $1\leq i\leq 100$. Each
classifier is fairly compared against the Bayes classifier via the distance 
\begin{equation}
d_{j}\!=\!\left\vert \bar{m}_{j}\!-\!m_{0}\right\vert \widehat{\sigma }%
_{m_{j}}\!+\!\left\vert \bar{q}_{j}\!-\!q_{0}\right\vert \widehat{\sigma }%
_{q_{j}},\;j\!\in \!\{\text{SP-SVM},\text{EEL-SVM},pin\text{-SVM},\overline{%
pin}\text{-SVM},\text{C-SVM}\},  \label{dist-def}
\end{equation}%
where $\bar{m}_{j}$ $(\bar{q}_{j})$ and $\widehat{\sigma }_{m_{j}}$ $(%
\widehat{\sigma }_{q_{j}})$ are respectively, the mean and standard
deviation estimates of $m_{j}$ $(q_{j})$ based on the $100$ point estimates.
Our results are reported in Table~\ref{denoiseresults} where we observe no
clear ranking among the methods under study for non-contaminated data,
though Section~\ref{num sec: sim data - with noise} shows a clear pattern
when data contamination is introduced.

\begin{table}[H]
\begin{center}
\begin{tabular}{crrrrr}
\hline
& C-SVM & $pin$-SVM & $\overline{pin}$-SVM & SP-SVM & EEL-SVM \\ 
\hline
$N=100$ & 0.3763 & 0.1132 & 0.1809 & \textbf{0.1014} & 0.3477 \\ 
$N=200$ & \textbf{0.0185} & 0.0337 & 0.0349 & 0.2166 & 0.0397 \\ \hline
\end{tabular}
\end{center}
\caption{Distance \eqref{dist-def} between various SVM classifiers and Bayes
classifier for non-contaminated synthetic data. The lowest distance for each
row is in bold.}
\label{denoiseresults}
\end{table}

\subsubsection{Synthetic Contaminated Data\label{num sec: sim data - with
noise}}

We next investigate how robust the five SVM classifiers are. This is
achieved by contaminating a percentage $r\in \lbrack 0,1]$ of the synthetic
data generated in Section~\ref{num sec: sim data - no noise}. Data
contamination is produced by generating random variates around a `central'
point from the `true' separation hyper-plane; without loss of generality,
the focal point is $(0,0)$. The contaminated data points are generated from
three elliptical distributions centered at $(0,0)$, namely, a bivariate
Normal $\mathcal{N}(\mathbf{0},\boldsymbol{\Sigma }_{c})$ and two bivariate
Student's $t(\mathbf{0},\boldsymbol{\Sigma }_{c},g)$ with $g\in \{5,1\}$
degrees of freedom and a covariance matrix given by 
\begin{equation}
\boldsymbol{\Sigma }_{c}=%
\begin{bmatrix}
1 & -0.8 \\ 
-0.8 & 1%
\end{bmatrix}%
.  \notag
\end{equation}%
The equal chance of labeling the response variable ensures even
contamination on both sides of the linear separation line; moreover, the
negative correlation $-0.8$ is chosen on purpose so that the DU becomes more
pronounced.

\begin{figure}[H]
\centering
\subfigure[Normal distribution]{
		\includegraphics[width=0.47\textwidth]{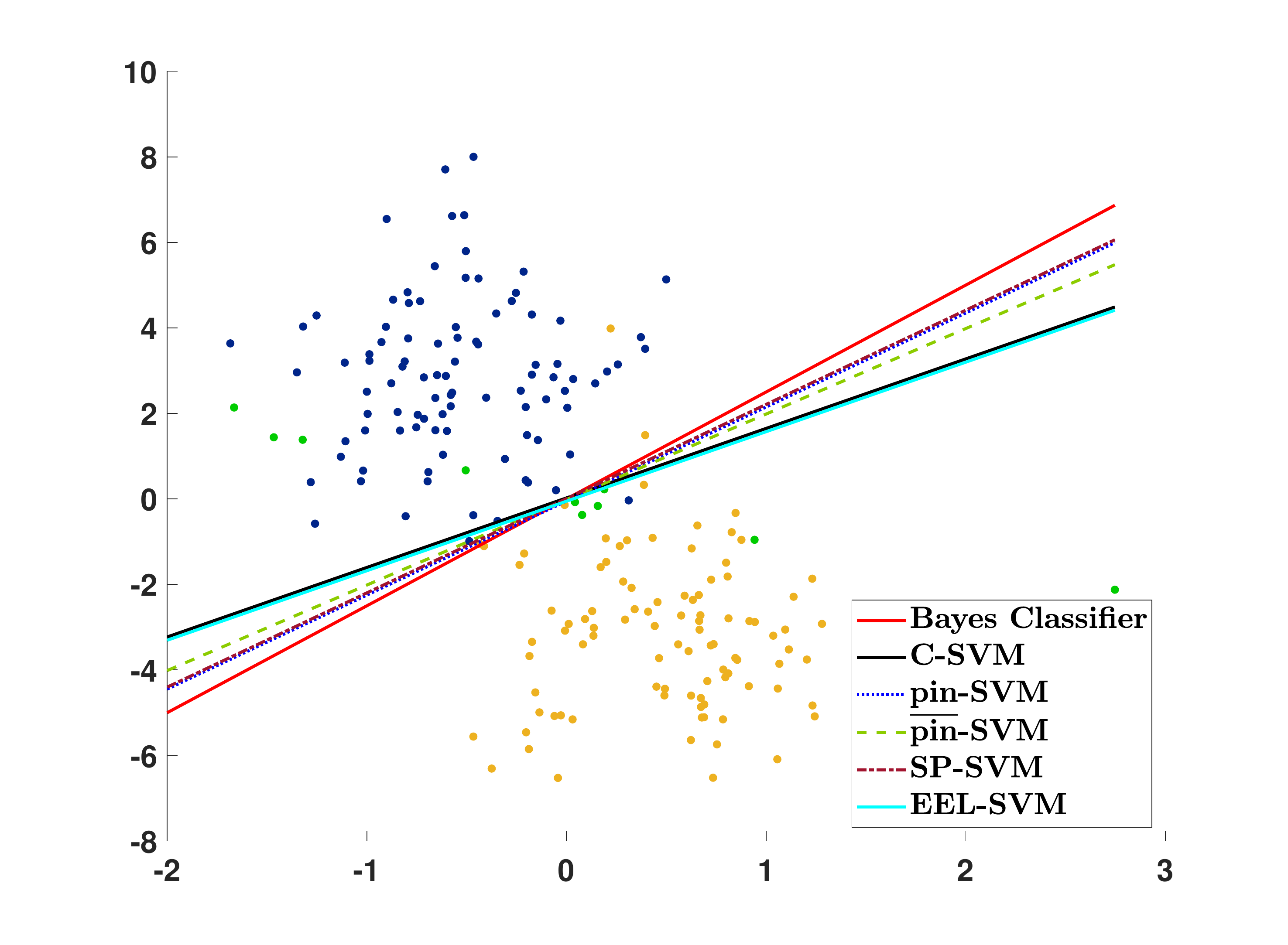}}  
\subfigure[Student's $t$ distribution with 5 degrees of freedom]{
		\includegraphics[width=0.47\textwidth]{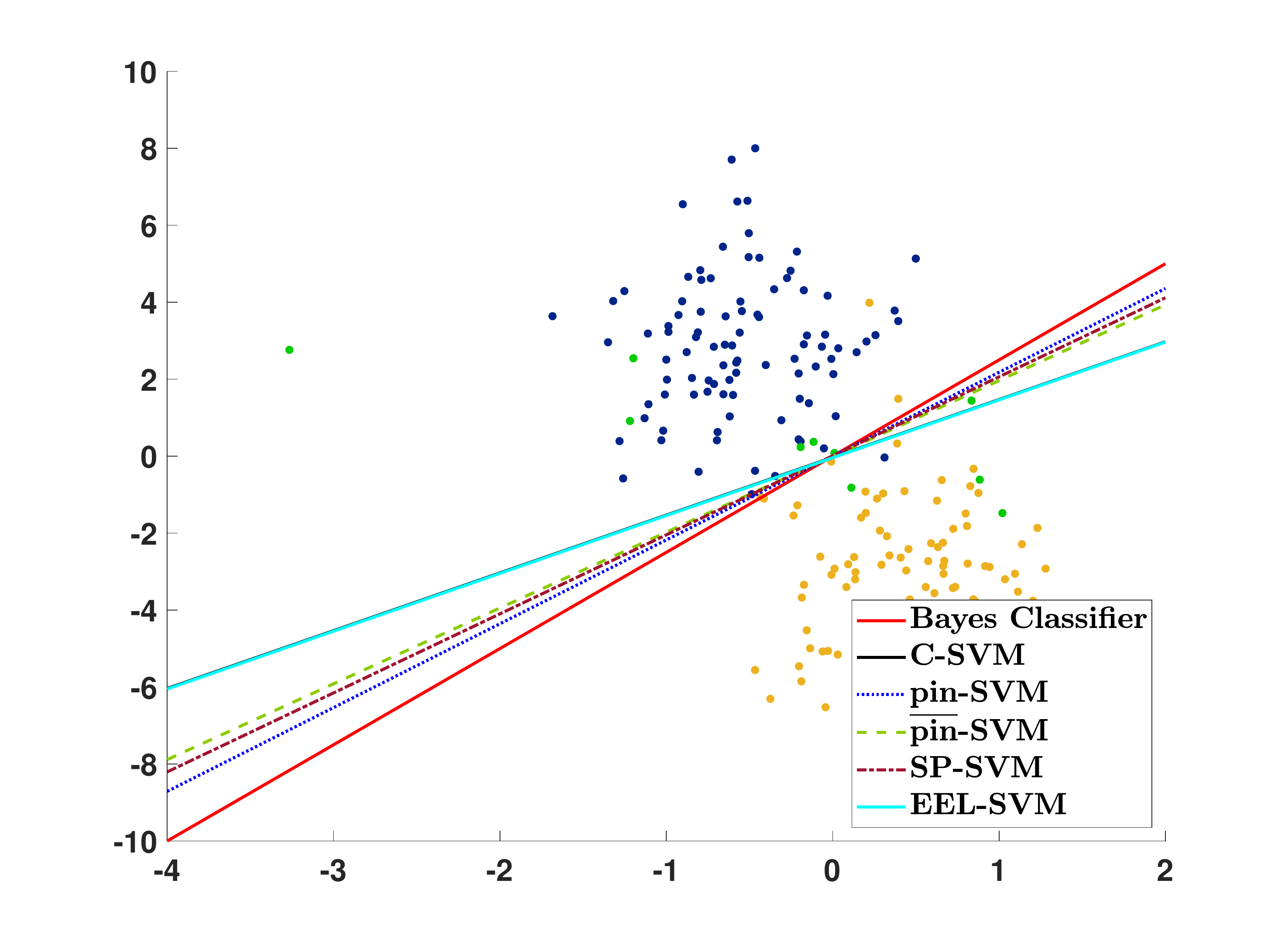}}\newline
\subfigure[Student's $t$ distribution with 1 degree of freedom]{
		\includegraphics[width=0.47\textwidth]{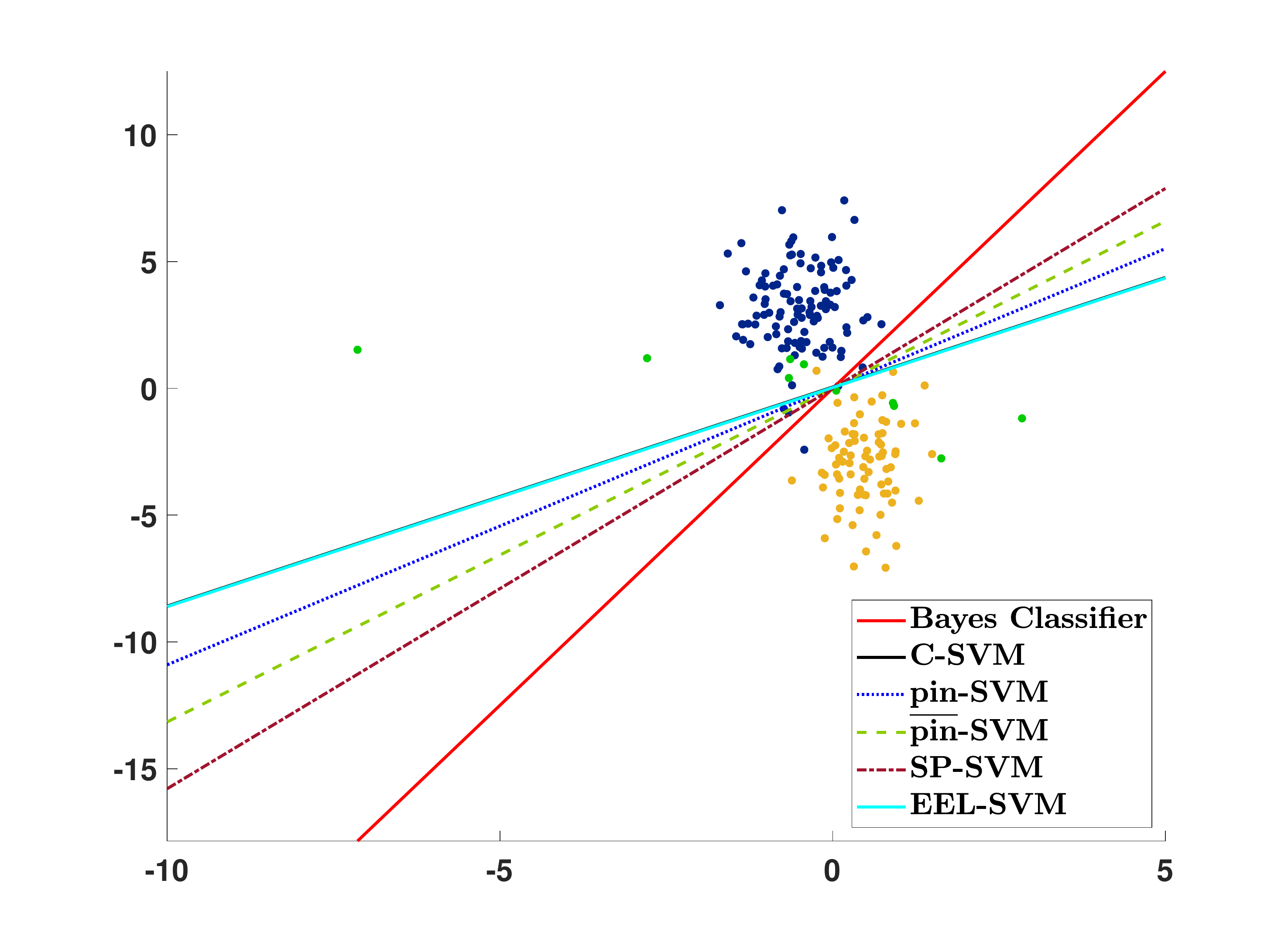}} 
\caption{Classification boundaries for five SVM classifiers if DU is induced by
(a) Normal distribution, (b) Student's $t$ distribution with 5 degrees of
freedom and (c) Student's $t$ distribution with 1 degree of freedom.}
\label{fig:syntheticdata}
\end{figure}

The contaminated data are plotted in Figure~\ref{fig:syntheticdata} together
with their estimated decision lines. The three scatter plots visualize the
realization of a sample of size $N=200$ that is contaminated with $r=5\%$,
and the contaminated points appear in green. If a decision rule is close to
the `true' decision rule given by the red solid line, then we could say that
its corresponding SVM classifier is more resilient to contamination, i.e is
more robust. Note that C-SVM and EEL-SVM overlap in Figure~\ref%
{fig:syntheticdata} because the contaminated data points become outliers
that are too extreme, even for EEL-SVM, and thus, this finds the same
decision rule as C-SVM. Figure~\ref{fig:syntheticdata}~(a) shows that most
of the classifiers, except C-SVM and EEL-SVM, are close to the `true'
classifier when a low level of DU is present due to the light-tailed
Gaussian contamination; the same behavior is observed in Figure~\ref%
{fig:syntheticdata}~(b), where a medium level of DU is present due to
Student's $t$ with 5 degrees of freedom contamination. Recall that the lower
the number of degrees of freedom is, the more heavy-tailed Student's $t$
distribution is; therefore, Figure~\ref{fig:syntheticdata}~(c) illustrates
the effect of a high level of DU, case in which SP-SVM seems to be the most
robust classifier.

The scatter plots in Figure~\ref{fig:syntheticdata} explain the
contamination mechanism, though these pictorial representations may be
misleading due to the sampling error, since Figure~\ref{fig:syntheticdata}
relies on a single random sample. Therefore, we repeat the same exercise $100
$ times in order to properly compare the classifiers in Tables~\ref%
{noiseresults} and \ref{computationaltimes}. Moreover, for each sample, we
conduct $10$-fold cross-validation to tune the additional parameters, and we
then compute the linear decision rule. The performance is measured via the
distance \eqref{dist-def}, and the summary of this analysis is provided in
Table~\ref{noiseresults} for samples of size $N\in \{100,200\}$ and a
contamination ratio $r\in \{0.05,0.10\}$. Note that the tuning parameters
are calibrated as in Section~\ref{num sec: sim data - no noise}, except
EEL-SVM, where the parameter space is enlarged (due to data contamination)
as follows: 
\begin{equation*}
\mathcal{A}_{EEL}:=\left\{ 
\begin{array}{cc}
\{0,0.01,\dots ,0.05\} & \text{if }r=0.05; \\ 
\{0,0.01,\dots ,0.10\} & \text{if }r=0.10.%
\end{array}%
\right. 
\end{equation*}%
EEL-SVM requires a larger parameter space when DU is more pronounced, i.e. $%
r=0.10$, but even in this extreme case the cardinality of $\mathcal{A}_{EEL}$
is not larger than the cardinality of any other parameter space. That is, we
do not favor EEL-SVM in the implementation phase.%

\begin{table}[H]
\begin{center}
\begin{tabular}{crrrr}
\hline
& \multicolumn{4}{c}{Normal distribution} \\ \hline
& \multicolumn{2}{c}{$r = 0.05$} & \multicolumn{2}{c}{$r = 0.10$} \\ 
& $N = 100$ & $N = 200$ & $N = 100$ & $N = 200$ \\ \hline
C-SVM & 0.8397 & 0.6623 & 1.1455 & 0.8675 \\ 
$pin$-SVM & \textbf{0.0704} & \textbf{0.1880} & \textbf{0.2955} & 0.2996 \\ 
$\overline{pin}$-SVM & 0.3073 & 0.3984 & 0.7334 & 0.4611 \\ 
SP-SVM & 0.2305 & 0.2994 & 0.5938 & \textbf{0.2813} \\ 
EEL-SVM & 0.8431 & 0.6788 & 1.1682 & 0.8785 \\ \hline\hline
& \multicolumn{4}{c}{Student's $t$ distribution (5 degrees of freedom)} \\ 
\hline
& \multicolumn{2}{c}{$r = 0.05$} & \multicolumn{2}{c}{$r = 0.10$} \\ 
& $N = 100$ & $N = 200$ & $N = 100$ & $N = 200$ \\ \hline
C-SVM & 1.1520 & 0.6795 & 1.4983 & 0.9863 \\ 
$pin$-SVM & \textbf{0.2966} & \textbf{0.1710} & \textbf{0.6861} & \textbf{\
0.3322} \\ 
$\overline{pin}$-SVM & {\ 0.5929} & {0.3410} & {0.9491} & {0.6492} \\ 
SP-SVM & 0.7405 & 0.3754 & 0.8801 & 0.4539 \\ 
EEL-SVM & 1.1560 & 0.6895 & 1.5077 & 1.0025 \\ \hline\hline
& \multicolumn{4}{c}{Student's $t$ distribution (1 degree of freedom)} \\ 
\hline
& \multicolumn{2}{c}{$r = 0.05$} & \multicolumn{2}{c}{$r = 0.10$} \\ 
& $N = 100$ & $N = 200$ & $N = 100$ & $N = 200$ \\ \hline
C-SVM & 1.8189 & 2.0358 & 2.6941 & 3.2549 \\ 
$pin$-SVM & \textbf{1.1983} & 1.6466 & 2.0853 & 2.1472 \\ 
$\overline{pin}$-SVM & {\ 1.4458} & {1.3479} & {1.9612} & {2.4647} \\ 
SP-SVM & 1.2384 & \textbf{1.2675} & \textbf{1.8261} & \textbf{1.8463} \\ 
EEL-SVM & 1.8558 & 2.0280 & 2.6788 & 3.2757 \\ \hline
\end{tabular}
\end{center}
\caption{Distance \eqref{dist-def} between various SVM classifiers and Bayes
classifier for contaminated synthetic data. The lowest distance for each row
is in bold.}
\label{noiseresults}
\end{table}

Table~\ref{noiseresults} shows that the performance of any classifier
deteriorates when the level of DU increases; moreover, the distance from the
Bayes classifier increases with the contamination ratio $r$. The overall
performances of SP-SVM and $pin$-SVM are superior to all other three
competitors, whereas $\overline{pin}$-SVM appears to be competitive in just
a few cases and C-SVM and EEL-SVM have a similar low performance. SP-SVM is
by far the most robust classifier when DU is more pronounced, which has been
observed in Figure~\ref{fig:syntheticdata}~(c) but for a single sample.

\begin{table}[H]
\begin{center}
\begin{tabular}{crrrr}
\hline
& \multicolumn{4}{c}{Normal distribution} \\ \hline
& \multicolumn{2}{c}{$r = 0.05$} & \multicolumn{2}{c}{$r = 0.10$} \\ 
& $N = 100$ & $N = 200$ & $N = 100$ & $N = 200$ \\ \hline
$pin$-SVM & 15.5138 & 33.9715 & 19.2942 & 39.8558 \\ 
$\overline{pin}$-SVM & 30.1658 & 22.9012 & 36.6373 & 32.7994 \\ 
SP-SVM & 7.6634 & 15.3068 & 9.2248 & 16.7844 \\ 
EEL-SVM & \textbf{6.6263} & \textbf{14.0950} & \textbf{8.6631} & \textbf{%
16.1476} \\ \hline\hline
& \multicolumn{4}{c}{Student's $t$ distribution (5 degrees of freedom)} \\ 
\hline
& \multicolumn{2}{c}{$r = 0.05$} & \multicolumn{2}{c}{$r = 0.10$} \\ 
& $N = 100$ & $N = 200$ & $N = 100$ & $N = 200$ \\ \hline
$pin$-SVM & 12.6893 & 41.3835 & 17.0174 & 44.2939 \\ 
$\overline{pin}$-SVM & 42.1007 & 26.7222 & 61.5340 & 27.6636 \\ 
SP-SVM & 6.1618 & 19.6716 & 7.7810 & 17.8432 \\ 
EEL-SVM & \textbf{5.6473} & \textbf{17.2680} & \textbf{7.7076} & \textbf{%
17.6688} \\ \hline\hline
& \multicolumn{4}{c}{Student's $t$ distribution (1 degree of freedom)} \\ 
\hline
& \multicolumn{2}{c}{$r = 0.05$} & \multicolumn{2}{c}{$r = 0.10$} \\ 
& $N = 100$ & $N = 200$ & $N = 100$ & $N = 200$ \\ \hline
$pin$-SVM & 11.4355 & 34.5926 & 15.0441 & 41.0832 \\ 
$\overline{pin}$-SVM & 25.4353 & 42.4561 & 45.9922 & 62.0633 \\ 
SP-SVM & 5.9118 & 15.5805 & 7.3598 & 17.8881 \\ 
EEL-SVM & \textbf{5.1893} & \textbf{14.5320} & \textbf{7.1204} & \textbf{%
16.3805} \\ \hline
\end{tabular}
\end{center}
\caption{Computational time ratios of SVM classifiers compared to C-SVM
classifier for contaminated synthetic data. The lowest computational time
ratio for each row is in bold.}
\label{computationaltimes}
\end{table}

We conclude our comparison by looking into the computational time ratios
(with C-SVM being the baseline reference) that are reported in Table~\ref%
{computationaltimes}. In particular, this provides the computational times
after tuning each model, i.e. the training computational time, which is a
standard and fair reporting when one would expect high computational times
when tuning more model hyper-parameters. EEL-SVM requires the lowest
computational effort, though it is very close to SP-SVM, while $pin$-SVM is
consistently slower and $\overline{pin}$-SVM is by far the method with the
largest computational time. These observations are not surprising because $%
\overline{pin}$-SVM relies on a non-convex (DCA) algorithm that has
scalability issues; on the contrary, $pin$-SVM, SP-SVM and EEL-SVM are
solved via convex LCQP of the same dimension, though EEL-SVM and $pin$-SVM
are, respectively, the most and least sparse.

\subsection{Real Data Analysis\label{num sec: real data}}

We now compare the classification accuracy for some real-life data and rank
all six SVM classifiers, including Ramp-KSVCR. Ten well-known real-world
datasets are chosen in this section, which can be retrieved from the UCI
depository\footnote{see \url{https://archive.ics.uci.edu/ml/index.php}} and LIBSVM depository\footnote{see \url{https://www.csie.ntu.edu.tw/~cjlin/libsvmtools/datasets/}}; a summary is
given in Table~\ref{datadescription}. It should be noted that all datasets
have features rescaled to $[-1,1]$. Moreover, the analysis is carried out
over the original and contaminated data. DU is introduced via the MATLAB
R2019a function \texttt{awgn} with different \emph{Signal Noise Ratios (SNR)}%
; perturbations are separately introduced $10$ times for each dataset before
training, and the average classification accuracy is reported so that the
sampling error is alleviated.

\begin{table}[H]
\begin{center}
\begin{tabular}{rlrrr}
\hline
& Data & Number of & Training & Testing \\ 
&  & features & sample size & sample size \\ \hline
(I) & Fourclass & 2 & 580 & 282 \\ 
(II) & Diabetes & 8 & 520 & 248 \\ 
(III) & Breast cancer & 10 & 460 & 223 \\ 
(IV) & Australian & 14 & 470 & 220 \\ 
(V) & Statlog & 13 & 180 & 90 \\ 
(VI) & Customer & 7 & 300 & 140 \\ 
(VII) & Trial & 17 & 520 & 252 \\ 
(VIII) & Banknote & 4 & 920 & 452 \\ 
(IX) & {A3a} & {123} & {3,185} & {29,376} \\ 
(X) & {Mushroom} & {\ 112} & {2,000} & {6,124} \\ \hline
\end{tabular}
\end{center}
\caption{Summary of the ten UCI datasets.}
\label{datadescription}
\end{table}

The data are randomly partitioned into the training and testing sets, as
described in Table~\ref{datadescription}. All SVM methods rely on the \emph{%
Radial Basis Function (RBF)} kernel chosen to overcome the lack of linearity
in the data. As before, SP-SVM methodology assumes that the feature with the
largest standard deviation is the one mostly affected by DU. All
hyper-parameters are tuned via a $10$-fold cross-validation; the kernel
parameter $\gamma $ and penalty parameter $C$ are tuned by allowing $\gamma
,C\in \big\{2^{-9},2^{-8},\dots ,2^{8},2^{9}\big\}$ for C-SVM, while $C\in %
\big\{2^{-5},2^{-3},2^{-1},2^{0},2^{1},2^{3},2^{5}\big\}$ and $\gamma \in %
\big\{2^{-7},2^{-5},2^{-3},2^{-1},2^{0},2^{1}\big\}$ are allowed for all
other classifiers. Note that C-SVM has fewer parameters than other
classifiers, and thus, $(\gamma ,C)$ are allowed more values in the tuning
process, so that all classifiers are treated as equally as possible. Note
that Ramp-KSVCR has two penalty parameters $C_{1},C_{2}$, an insensitivity
parameter $\epsilon $ and two additional model parameters $s$ and $t$; the
penalty parameters satisfy $C_{1}=C_{2}$, as in the original paper. The
other parameters are tuned as follows:

\begin{itemize}
\item SP-SVM: $\alpha \in \mathcal{A}_{SP}=\{0.50,0.51,\dots
,0.56,0.58,0.60\}$;

\item EEL-SVM: $\alpha \in \mathcal{A}_{EEL}=\{0,0.05,0.10,\dots ,0.3\}$;

\item $pin$-SVM: $\tau \in \mathcal{T}_{pin}=\{0.1,0.2,\dots ,0.8,1\}$;

\item $\overline{pin}$-SVM$: s\in \{0.01, 0.1, 0.3, 0.5, 0.7, 1\}$ and $\tau
= 0.5$;

\item Ramp-KSVCR: $\epsilon \in \{0.1, 0.2, 0.3\}, t\in \{1,3, 5\}, s=-1$.
\end{itemize}

\noindent A final note is that a computational budget of around $370$ parameter
combinations is imposed on all cases, except EEL-SVM where $294$
combinations are considered.

\begin{table}[H]
{\footnotesize \renewcommand{\arraystretch}{0.95}  
\hskip+1.25cm  
\begin{tabular}{ccrrrrrr}
\hline
Data & SNR & C- & $pin$- & $\overline{pin}$- & SP- & EEL- & Ramp- \\ 
&  & SVM & SVM & SVM & SVM & SVM & KSVCR \\ \hline
\multirow{4}{*}{(I)} & NA & 99.29\% & 99.29\% & \textbf{{\ {\ 99.65\%}}} & 
\textbf{99.65\%} & \textbf{99.65\%} & {\ 92.20\%} \\ \cline{2-8}
& 10 & 99.65\% & 99.65\% & {\ 99.65\%} & 99.61\% & \textbf{99.75\%} & {\
91.95\%} \\ 
& 5 & 99.65\% & 99.65\% & {\ 99.65\%} & 99.54\% & \textbf{99.72\%} & {\
91.67\%} \\ 
& 1 & 99.54\% & 99.61\% & {\ 99.61\%} & 99.50\% & \textbf{99.65\%} & {\
91.84\%} \\ \hline
\multirow{4}{*}{(II)} & NA & 77.02\% & 79.84\% & {\ 79.84\%} & \textbf{%
80.24\%} & 78.63\% &  \textbf{80.24\%} \\ \cline{2-8}
& 10 & 76.98\% & 76.49\% & {\ 78.10\%} & \textbf{79.64\%} & 77.26\% & {\
77.10\%} \\ 
& 5 & 76.69\% & 76.57\% & {\ 77.54\%} & 78.02\% & 76.45\% &  \textbf{79.48\%}
\\ 
& 1 & 76.49\% & 77.70\% & {\ 76.90\%} & 77.66\% & 74.96\% &  \textbf{79.44\%}
\\ \hline
\multirow{4}{*}{(III)} & NA & 93.72\% & 93.27\% & {\ 94.17\%} & 94.62\% & 
93.72\% &  \textbf{95.96\%} \\ \cline{2-8}
& 10 & 93.90\% & 94.75\% & {\ 93.86\%} & 93.32\% & 94.44\% &  \textbf{95.29\%%
} \\ 
& 5 & 93.86\% & 94.57\% & {\ 94.17\%} & 94.13\% & 94.08\% &  \textbf{94.80\%}
\\ 
& 1 & 93.81\% & 93.86\% & {\ 93.86\%} & 93.86\% & 94.04\% &  \textbf{95.11\%}
\\ \hline
\multirow{4}{*}{(IV)} & NA & 88.64\% & 88.18\% &  \textbf{89.55\%} & 88.18\%
& 89.09\% &  \textbf{89.55\%} \\ \cline{2-8}
& 10 & \textbf{85.82\%} & 85.23\% & {\ 85.77\%} & 85.45\% & 85.32\% & {\
83.82\%} \\ 
& 5 & 80.68\% & 80.50\% &  \textbf{\ 82.41\%} & 80.59\% & 78.45\% & {\
77.73\%} \\ 
& 1 & 76.59\% & 75.86\% & \textbf{\ 77.86\%} & 76.14\% & 76.23\% & {\ 75.77\%%
} \\ \hline
\multirow{4}{*}{(V)} & NA & 82.22\% & 82.22\% & {\ 82.22\%} & \textbf{83.33\%%
} & 78.89\% & {\ 80.00\%} \\ \cline{2-8}
& 10 & 80.22\% & 80.56\% & {\ 77.56\%} & 80.22\% & \textbf{82.44\%} & {\
81.00\%} \\ 
& 5 & 80.00\% & 79.33\% & {\ 79.89\%} & 79.33\% & \textbf{\ 81.33\%} & {\
77.00\%} \\ 
& 1 & 78.67\% & 78.22\% & {\ 80.00\%} & 78.44\% & 76.89\% &  \textbf{80.22\%}
\\ \hline
\multirow{4}{*}{(VI)} & NA & 92.14\% & 91.43\% & {\ 90.71\%} & 92.14\% & 
\textbf{92.86\%} & {\ 91.89\%} \\ \cline{2-8}
& 10 & 92.86\% & 92.50\% & {\ 92.50\%} & 92.71\% & \textbf{93.21\%} & {\
89.93\%} \\ 
& 5 & 92.93\% & 92.86\% & {\ 91.43\%} & \textbf{93.07\%} & \textbf{93.07\%}
& {\ 91.36\%} \\ 
& 1 & 92.57\% & \textbf{92.93\%} & {\ 91.57\%} & 92.57\% & 92.86\% & {\
90.79\%} \\ \hline
\multirow{4}{*}{(VII)} & NA & \textbf{\ 100.00\%} & \textbf{\ 100.00\%} &  
\textbf{\ 100.00\%} & \textbf{\ 100.00\%} & \textbf{\ 100.00\%} &  \textbf{\
100.00\%} \\ \cline{2-8}
& 10 & 99.56\% & \textbf{99.80\%} & {\ 99.33\%} & 99.76\% & 99.60\% & {\
99.48\%} \\ 
& 5 & 94.72\% & 94.60\% & {\ 94.44\%} & \textbf{\ 94.84\%} & 94.52\% & {\
93.97\%} \\ 
& 1 & 88.13\% & 88.13\% & {\ 85.99\%} & \textbf{\ 88.29\%} & 88.21\% & {\
86.98\%} \\ \hline
\multirow{4}{*}{(VIII)} & NA & \textbf{\ 100.00\%} & \textbf{\ 100.00\%} &  
\textbf{\ 100.00\%} & \textbf{\ 100.00\%} & \textbf{\ 100.00\%} &  \textbf{\
100.00\%} \\ \cline{2-8}
& 10 & 99.73\% & 99.71\% & {\ 99.80\%} & 99.76\% & \textbf{\ 99.89\%} & {\
99.85\%} \\ 
& 5 & \textbf{\ 99.38\%} & 99.18\% & {\ 99.05\%} & 99.25\% & 99.36\% & {\
99.05\%} \\ 
& 1 & \textbf{\ 97.94\%} & 97.92\% & {\ 97.61\%} & \textbf{\ 97.94\%} & 
97.83\% & {\ 96.75\%} \\ \hline
\multirow{4}{*}{(IX)} & NA & {\ 82.81\%} & {\ 83.20\%} & {\ 83.36\%} & {\
83.67\%} & {\ 81.71\%} &  \textbf{84.04\%} \\ \cline{2-8}
& 10 & {\ 80.50\%} & {\ 81.05\%} & {\ 80.57\%} & {\ 81.12\%} &  \textbf{\
81.15\%} & {\ 81.13\%} \\ 
& 5 &  \textbf{\ 78.57\%} & {\ 78.56\%} & {\ 77.94\%} & {\ 78.35\%} & {\
78.26\%} & {\ 78.00\%} \\ 
& 1 & {\ 76.34\%} &  \textbf{76.74\%} & {\ 75.94\%} & {\ 76.01\%} & {\
76.02\%} & {\ 76.25\%} \\ \hline
\multirow{4}{*}{(X)} & NA &  \textbf{99.87\%} &  \textbf{99.87\%} &  \textbf{%
99.87\%} &  \textbf{\ 99.87\%} &  \textbf{99.87\%} &  \textbf{99.87\%} \\ 
\cline{2-8}
& 10 & {\ 98.37\%} & {\ 98.84\%} &  \textbf{\ 99.38\%} & {\ 98.31\%} & {\
98.29\%} & {\ 99.03\%} \\ 
& 5 & {\ 93.02\%} & {\ 93.79\%} &  \textbf{\ 94.72\%} & {\ 93.05\%} & {\
92.92\%} & {\ 93.77\%} \\ 
& 1 & {\ 85.36\%} & {\ 85.63\%} &  \textbf{\ 85.82\%} & {\ 85.46\%} & {\
85.47\%} & {\ 85.55\%} \\ \hline
\multicolumn{2}{c}{Average rank} & 3.25 & 3.18 & 3.18 & \textbf{2.93} & 3.05
& 3.53 \\ \hline
\end{tabular}
}
\caption{Classification accuracy (in $\%$) of all SVM classifiers across all
datasets. The highest accuracies for each row are in bold. Each row signifies
the original data (reported as ``NA'', i.e. no contamination) or their
contaminated variants (with SNR values of $\{1,5,10\}$).}
\label{realdata-set}
\end{table}

Table~\ref{realdata-set} summarizes the classification performance for all
ten datasets for the original (non-contaminated) data and their contaminated
variants with various SNR values, where smaller SNR value means higher
degree of data contamination. EEL-SVM achieves the best performance in 14
out of 40 scenarios investigated, which is followed by Ramp-KSVCR that
performs best in 13 out of 40 scenarios; the other classifiers are ranked as
follows: SP-SVM (11/40), $\overline{pin}$-SVM (10/40), C-SVM (7/40) and $pin$%
-SVM (6/40). We also calculate the average ranks for each classifier by
looking at each row of Table~{\ref{realdata-set}} and rank each entry from 1
to 6 from lowest to highest accuracy; SP-SVM and EEL-SVM yield the best
average ranks among all classifiers and SP-SVM outperforms its competitors
via this criterion.

In summary, SP-SVM and EEL-SVM exhibit competitive advantage over their
competitors, with EEL-SVM being more sparse, though SP-SVM seems to be
performing slightly better.

\subsection{Interpretable Classifiers\label{fin_data_sec}}


The performance of SVM classifiers for real-life data has been analyzed in
Section~\ref{num sec: real data} without interpreting the decision rules so
that the presence of DU is better understood and the fairness of the
decision is assessed. We achieve that now by providing a granular analysis
for classification of the US mortgage lending data that are downloaded from
the \emph{Home Mortgage Disclosure Act (HMDA)} website (%
\url{https://ffiec.cfpb.gov}). Specifically, we collected the 2020 data for
two states, namely, \emph{Maine (ME)} 
and \emph{Vermont (VT)}, with a focus on \emph{subordinated lien} mortgages.



Subordinate-lien (`piggyback') loans are taken out at the same time as
first-lien mortgages on the same property by borrowers, mainly to avoid
paying mortgage insurance on the first-lien mortgage (due to the extra down
payment). 
\citet{eriksen.al.13} find evidence that borrowers with subordinate loans
have an increased by 62.7\% chance to default each month on primary loan.
Such borrowers may sequentially default on each loan since subordinate
lenders will not pursue foreclosure if the borrowers have insufficient
equity until at least housing markets start to recover. 
As noted by \citet{soyer.xu.10}, due to major costs from default for all involved parties, including mortgage lenders and mortgagors, the modelling, assessment, and management of mortgage default risk (e.g., \citealp{bhattacharya.al.19}) is a major concern for financial institutions and policy makers.  

Subordinate-lien loans are high-risk mortgages and we aim to classify the
instances as `loan originated' ($Y=+1$) or `application denied' ($Y=-1$)
using the available features. The HMDA data have numerous features and the
following representative ones are chosen: F1) loan amount, F2) loan to value
ratio (F1 divided by the `property\_value'), F3) percentage of minority
population to total population for tract, F4) percentage of tract median
family income compared to MSA (metropolitan statistical area) median family
income. Two categorical features are also considered, namely, F5) derived
sex and F6) age.

All categorical features are pre-processed via the standard one-hot encoding
procedure and all features are rescaled to $[-1,1]$ before training. A
random sampling is performed to extract the training and testing sets, so
the training set is twice as large as the testing set. The hyper-parameter
tuning of the three methods (C-SVM, SP-SVM, EEL-SVM)\ is performed via $10$%
-fold cross-validation using the hyper-parameter spaces in Section~\ref{num
sec: real data}. SP-SVM identifies F1 (loan amount) as the feature affected
by DU, which has the largest standard deviation at the same time. This is
not surprising since the loan amount has a massive impact on the mortgage
lending decision and this feature is heavily influenced by all other
features. Table \ref{denoiseresults--2} reports the details of the
training-testing splitting and the out-of-sample accuracy of the three SVM
methods. 
\begin{table}[tbp]
\begin{center}
\begin{tabular}{crrrrr}
\hline
Dataset & Total & Testing & C-SVM & SP-SVM & EEL-SVM \\ 
& sample size & sample size &  &  &  \\ \hline
ME & 4,226 & 1,396 & \textbf{70.77\%} & \textbf{70.77}\% & 70.13\% \\ 
VT & 1,948 & 648 & 90.74\% & \textbf{91.67}\% & 90.89\% \\ \hline
\end{tabular}
\end{center}
\caption{Summary of the HMDA datasets and their accuracy levels. The highest
accuracy for each row is in bold.}
\label{denoiseresults--2}
\end{table}
We observe that SP-SVM performs best in terms of accuracy, with C-SVM and
EEL-SVM relatively close. The next step is to interpret the classification
rule and explain the DU effect, but also evaluate the effect of an
automatized mortgage lending decision. The latter is measured by looking
into unfavorable decisions where the loan is denied, i.e. $Y=-1$.

The \emph{Equal Credit Opportunity Act (ECOA)} prohibits a creditor from
discriminating against any borrower on the basis of age, marital status,
race, religion, or sex, known as protected characteristics; such a
regulatory requirement is imposed not only in the US, but also similar ones
are in place in the EU, UK and elsewhere. Under ECOA, regulatory agencies assess
the lending decision fairness of lending institutions by comparing the
unfavorable decision ($Y=-1$) across different groups with given protected
characteristics. Our next analysis focuses on checking whether an
automatized lending decision could lead to unintentional discrimination,
known as \emph{disparate impact}. First, we look at the entry data and
provide evidence on whether or not the loan amount is massively different
across the applicants' gender at birth, income and racial structure in their
postal code, which would explain if DU is present or not. Second, we
evaluate the fairness of the lending decision obtained via SVM
classification and argue which SVM-based decision is more
compliant with such non-discrimination regulation (with respect to the sex attribute).

\begin{table}[H]
\begin{center}
\begin{tabular}{clrrr|rrr}
\hline
&  & \multicolumn{3}{c|}{Training data} & \multicolumn{3}{c}{Testing data}
\\ \hline
&  & MvsF & MvsJ & FvsJ & MvsF & MvsJ & FvsJ \\ \hline
\multirow{3}{*}{ME} & $Y \in \{-1, 1\}$ & 0.0853 & 0.0477 & \textbf{0.1330}
& 0.0969 & 0.1248 & \textbf{0.2096} \\ 
& $Y = -1$ & 0.0540 & 0.0727 & \textbf{0.1215} & 0.1203 & 0.1122 & \textbf{%
0.2313} \\ 
& $Y = 1$ & 0.1091 & 0.0381 & \textbf{0.1472} & 0.0889 & 0.1266 & \textbf{%
0.1986} \\ \hline
\multirow{3}{*}{VT} & $Y \in \{-1, 1\}$ & 0.0923 & 0.1009 & \textbf{0.1841}
& 0.1355 & 0.1399 & \textbf{0.2000} \\ 
& $Y = -1$ & 0.0866 & 0.0449 & \textbf{0.1208} & \textbf{0.3095} & 0.1429 & 
0.1667 \\ 
& $Y = 1$ & 0.1258 & 0.1579 & \textbf{0.2515} & 0.1235 & 0.1575 & \textbf{%
0.2108} \\ \hline
\end{tabular}
\end{center}
\caption{Kolmogorov--Smirnov distances in the loan amount distribution of
males versus females (MvsF), males versus joint (MvsJ) and females versus
joint (FvsJ). The largest distances for each row corresponding to training
and testing data are in bold.}
\label{tab:KS-finance}
\end{table}

Table \ref{tab:KS-finance} reports the \emph{Kolmogorov--Smirnov} distances
for loan amount samples of applicants based on gender characteristics. There
is overwhelming evidence that joint loan applications and female applicants
have very different loan amount distributions in both training and testing
data, though VT data exhibit the largest distance when comparing male and
female applicants with a favorable mortgage lending decision. This could be
explained by socio-economic disparities between males and females, though DU
plays a major role in this instance. Gender information in the HMDA data is
expected to have a self-selection bias, since applicants at risk are quite
unlikely to report gender information as they believe that the lending
decision would be influenced by that. Consequently, we removed a significant
portion of the data, i.e. examples for which the gender information is
unknown, which is clear evidence of self-selection bias in our entry data.

\begin{figure}[H]
\centering
\includegraphics[width=0.32\textwidth,height=5cm]{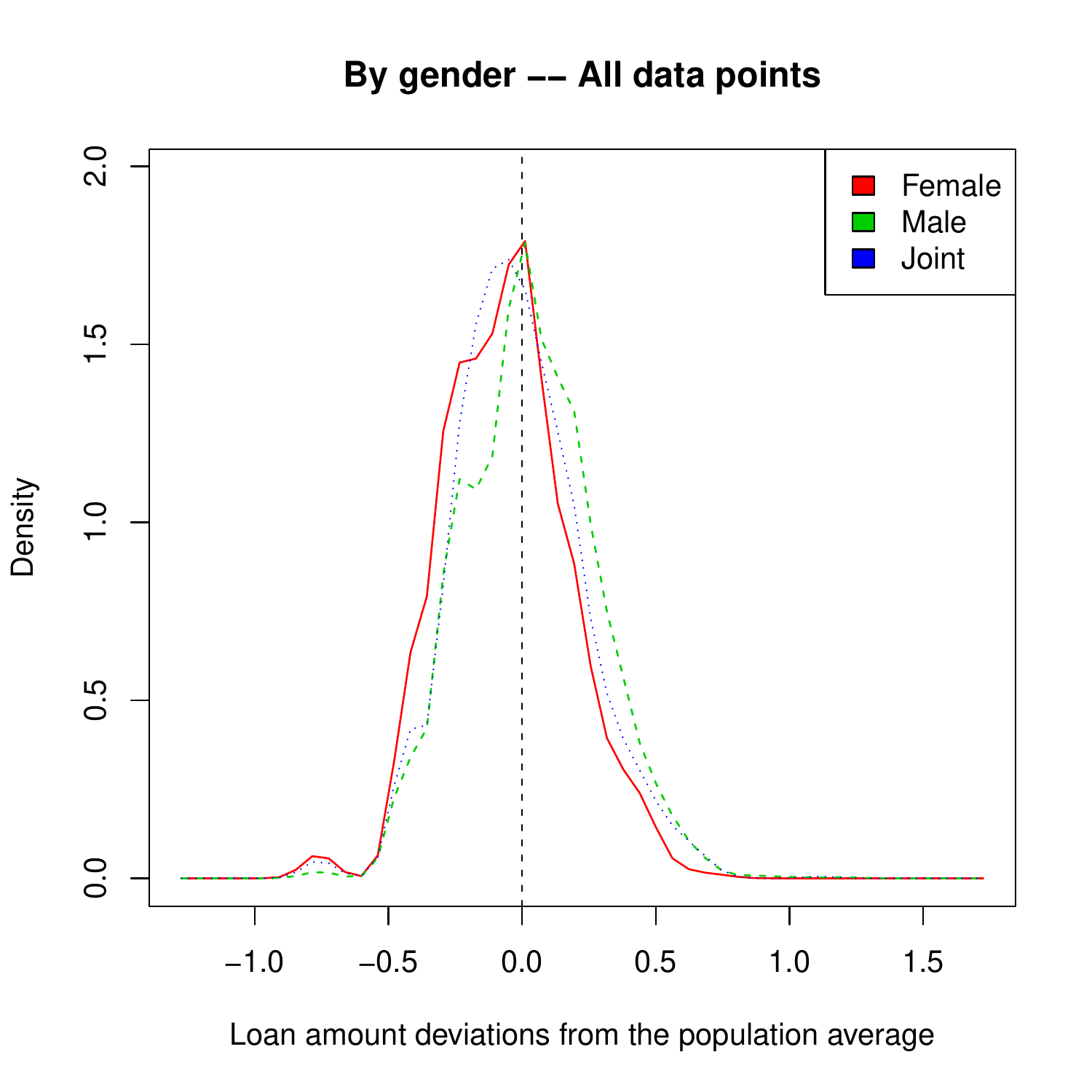} %
\includegraphics[width=0.32%
\textwidth,height=5cm]{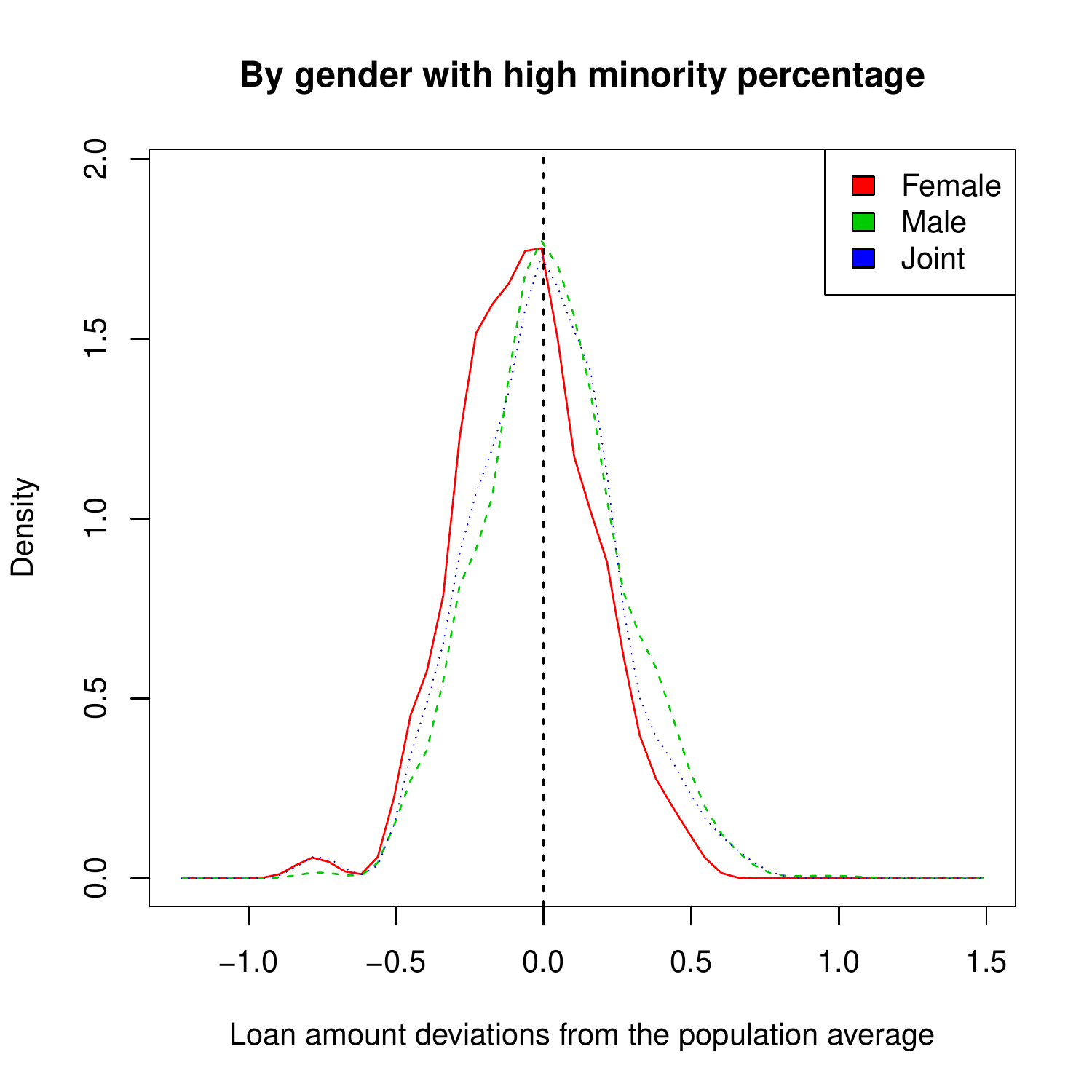} %
\includegraphics[width=0.32%
\textwidth,height=5cm]{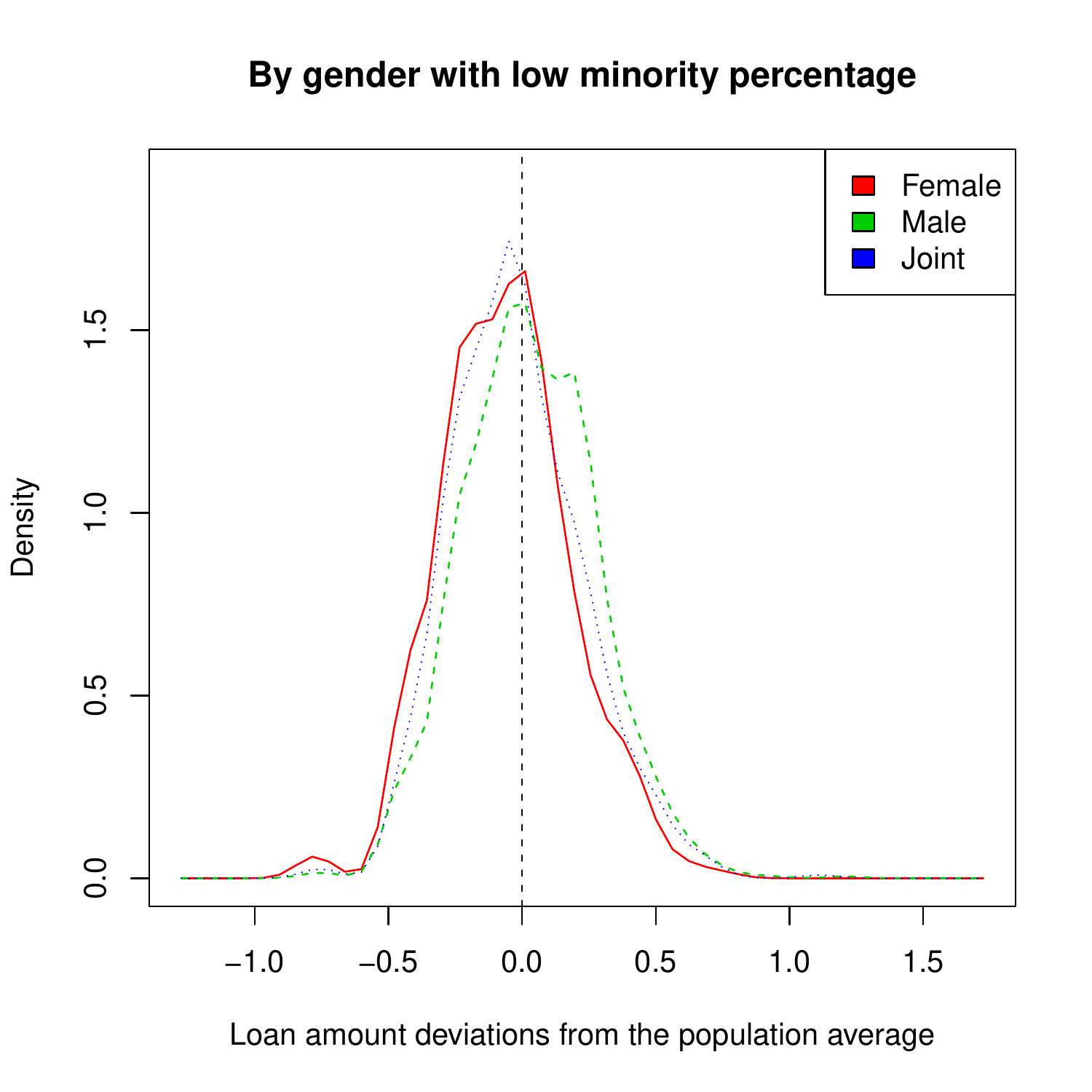}\newline
\hspace{2.1in}\includegraphics[width=0.32%
\textwidth,height=5cm]{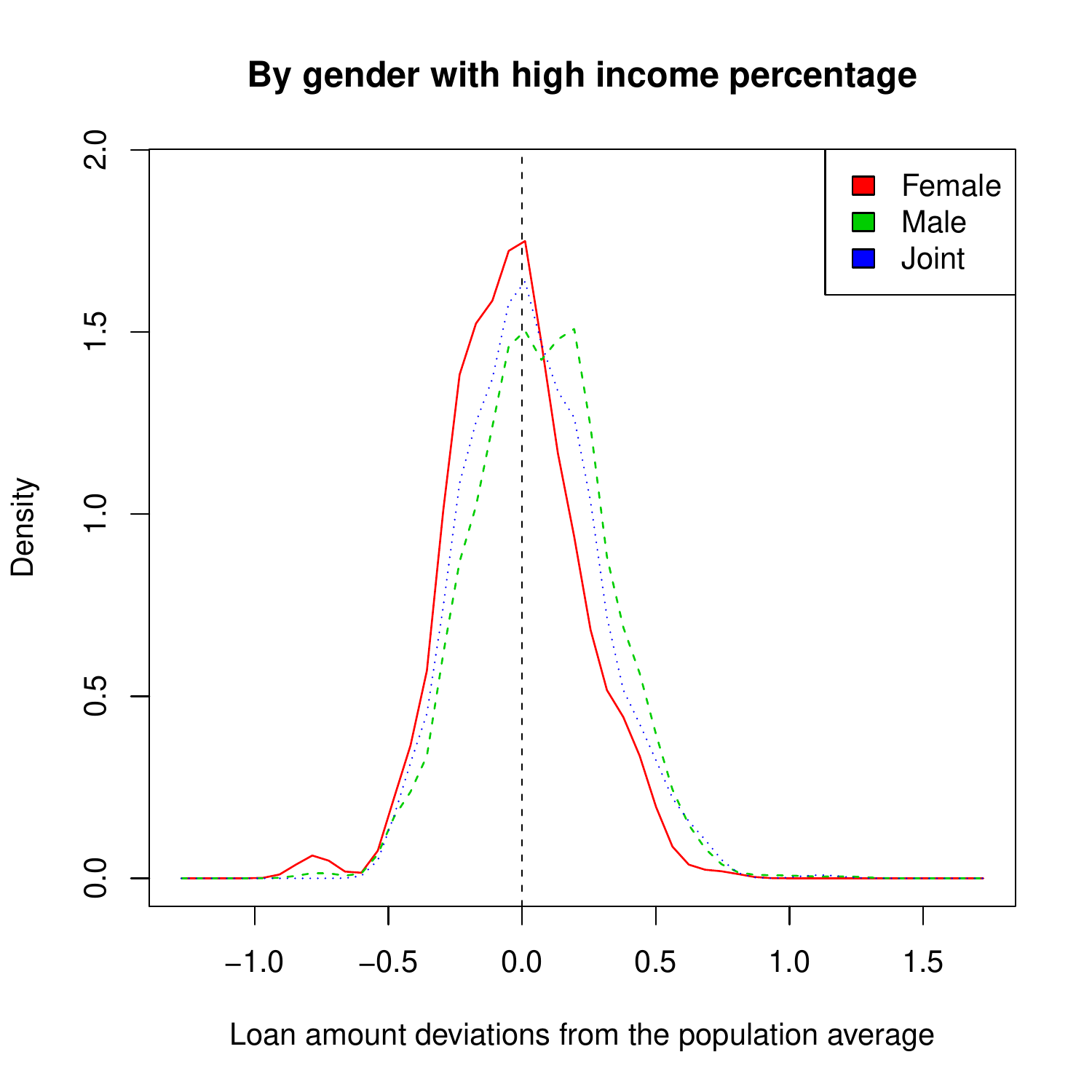} %
\includegraphics[width=0.32%
\textwidth,height=5cm]{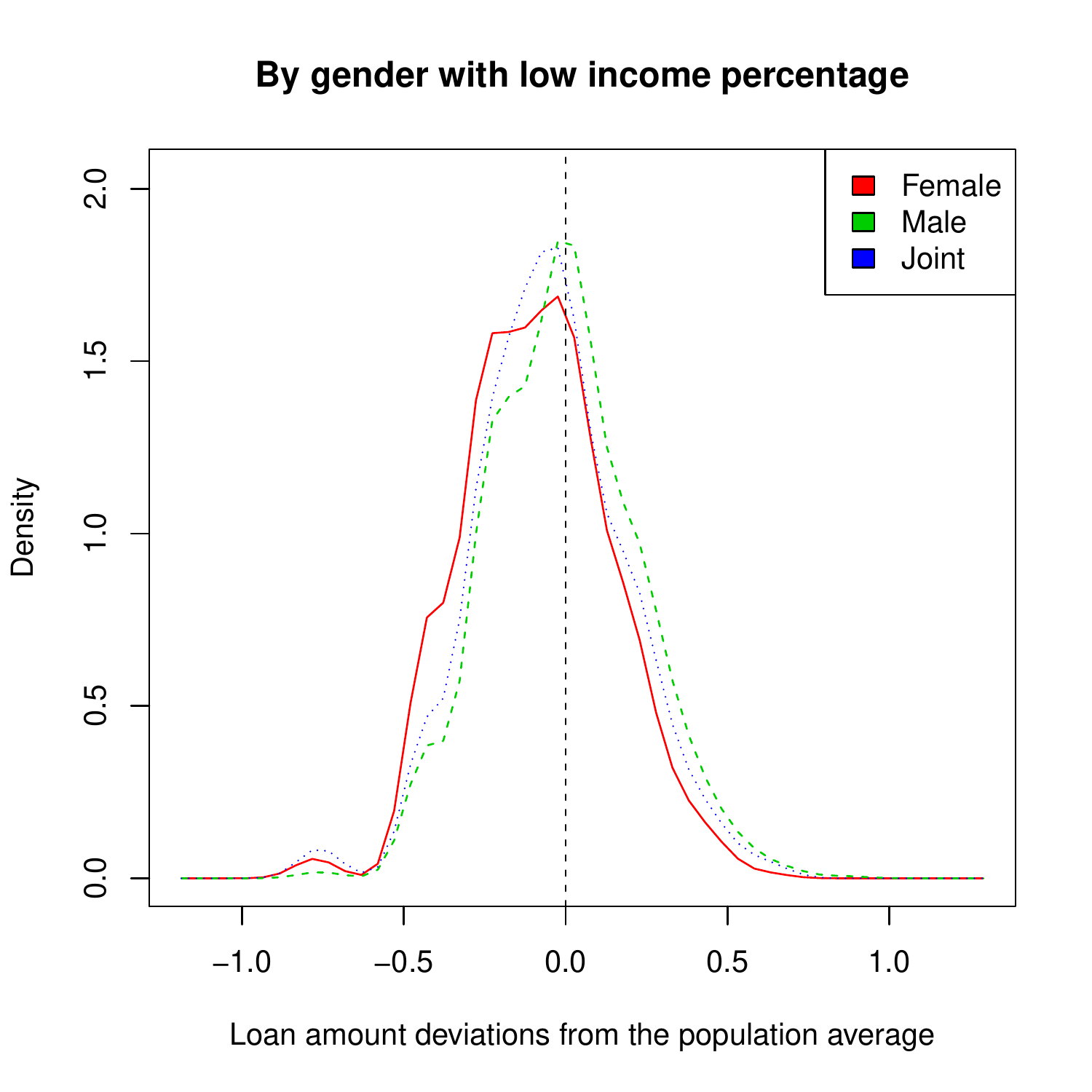} 
\caption{ME loan amount deviations from the population mean based on full data and sub-populations with low/high minority and low/high income percentages.}
\label{fig:HDMA_DU_ME}
\end{figure}

\begin{figure}[H]
\centering
\includegraphics[width=0.32\textwidth,height=5cm]{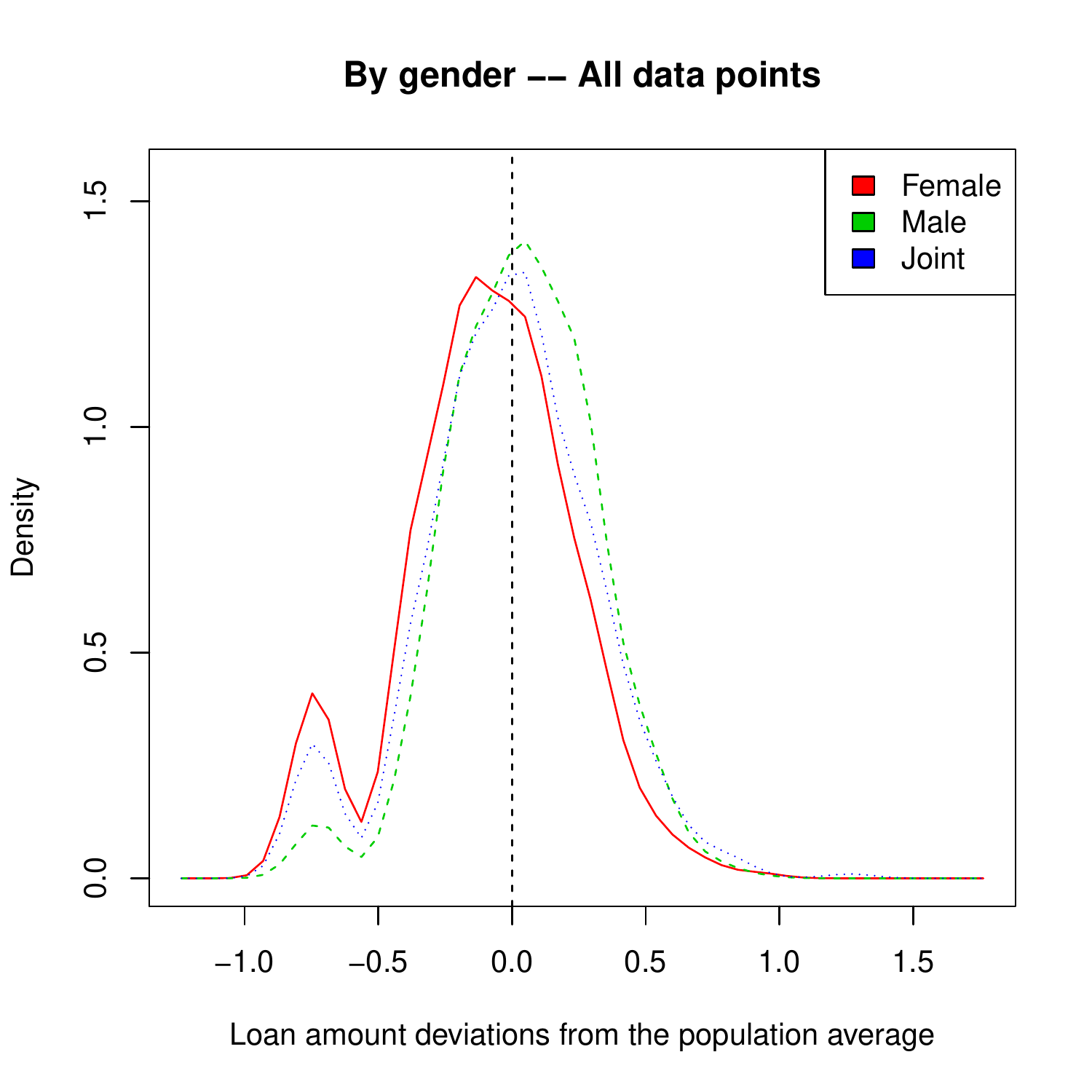} %
\includegraphics[width=0.32%
\textwidth,height=5cm]{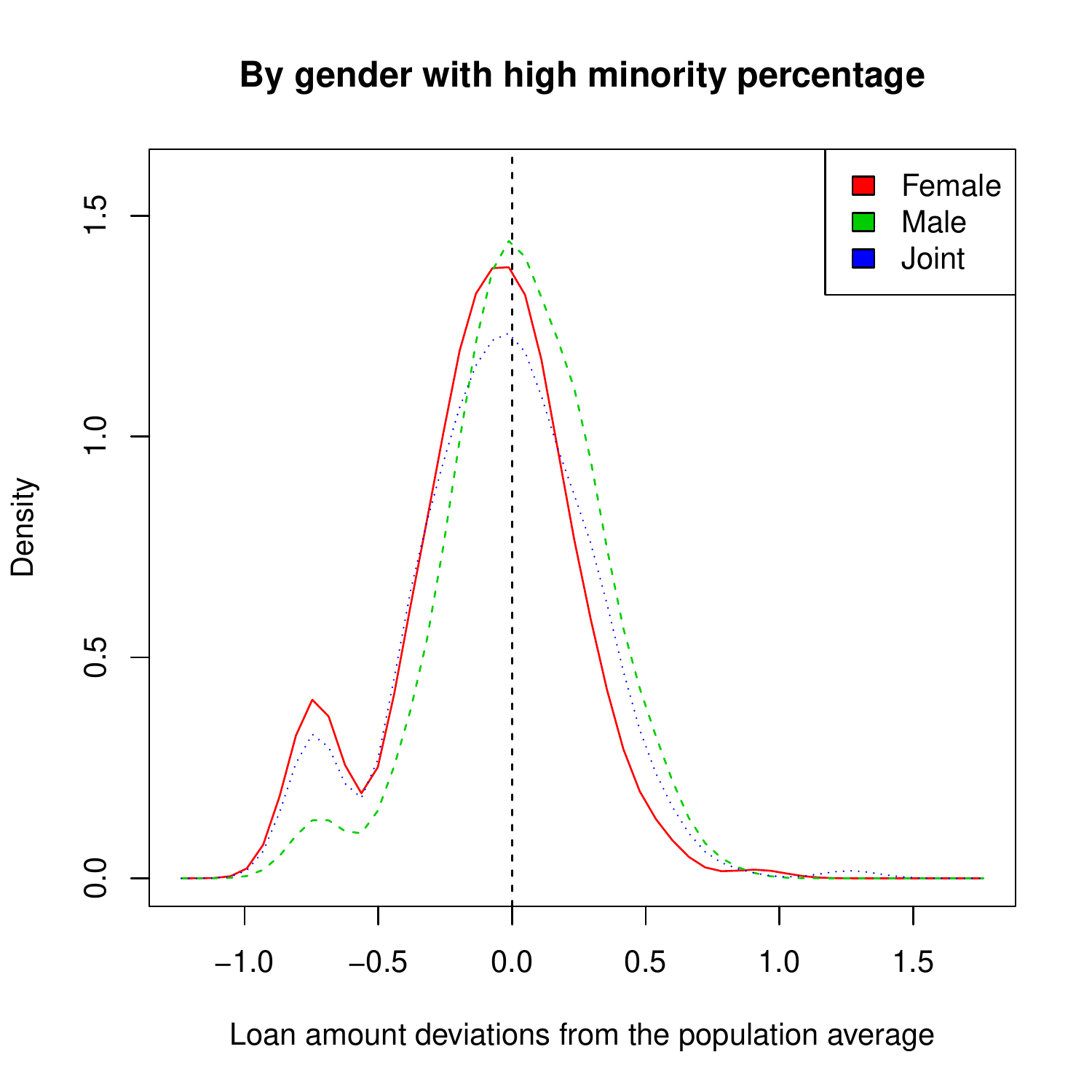} %
\includegraphics[width=0.32%
\textwidth,height=5cm]{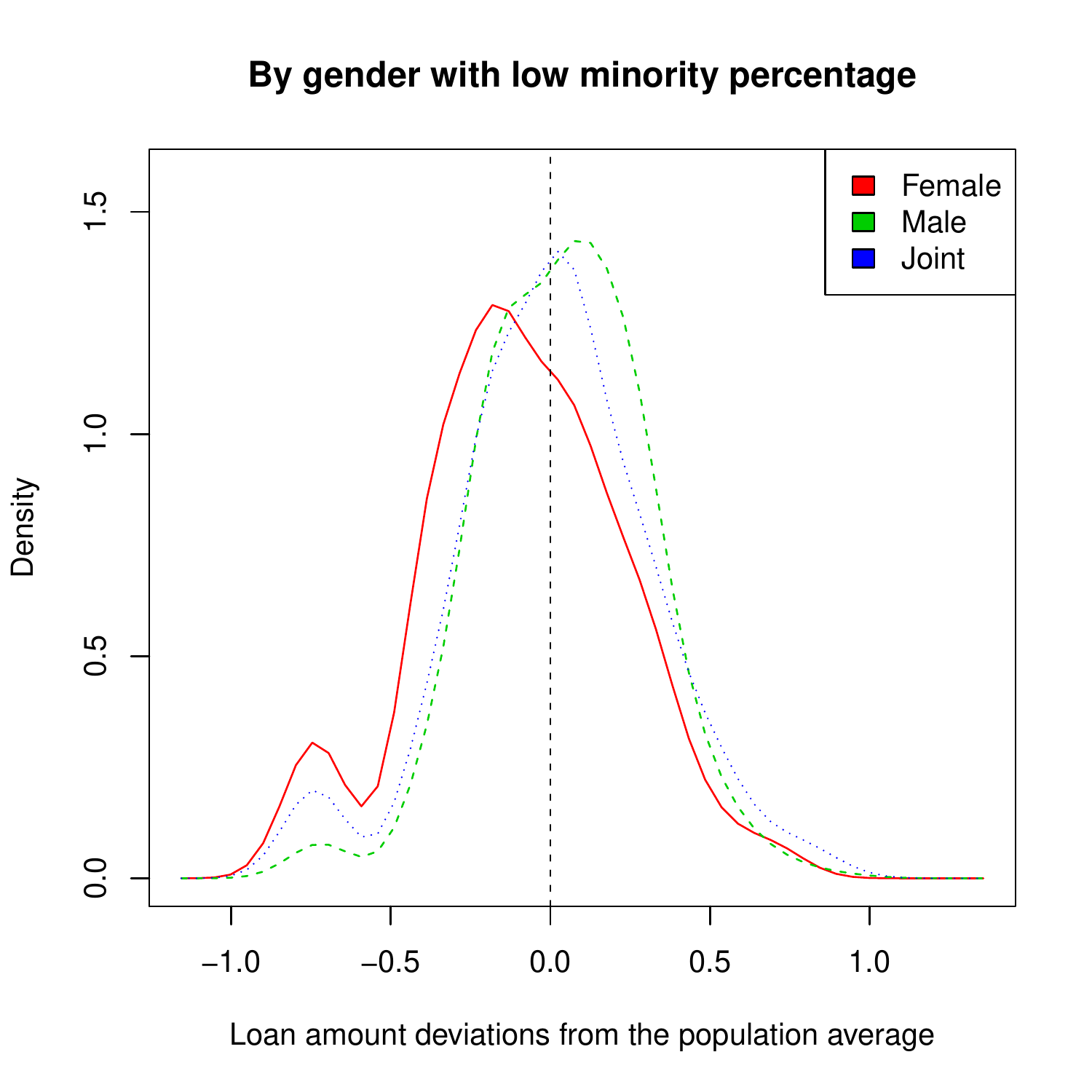}\newline
\hspace{2.1in}\includegraphics[width=0.32%
\textwidth,height=5cm]{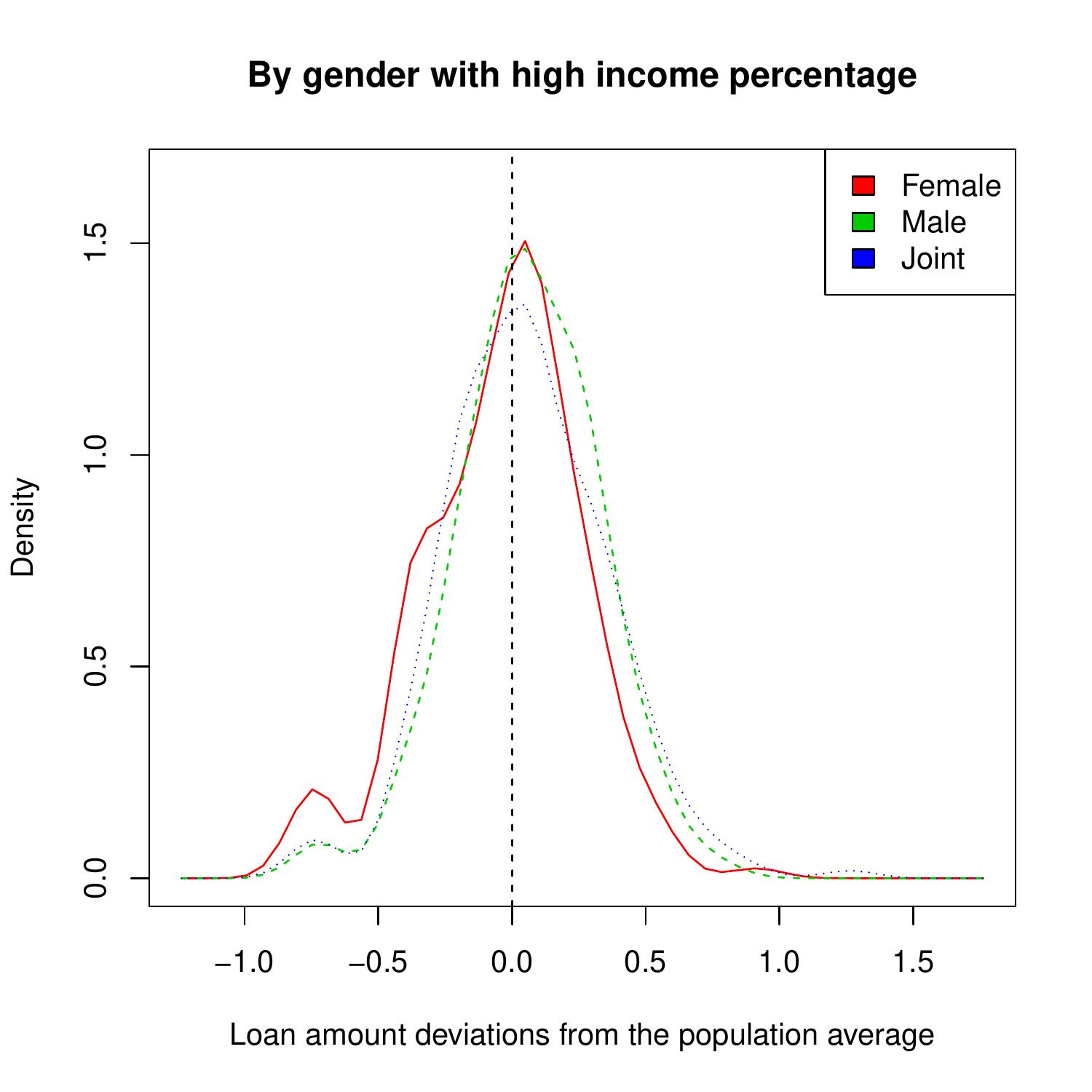} %
\includegraphics[width=0.32%
\textwidth,height=5cm]{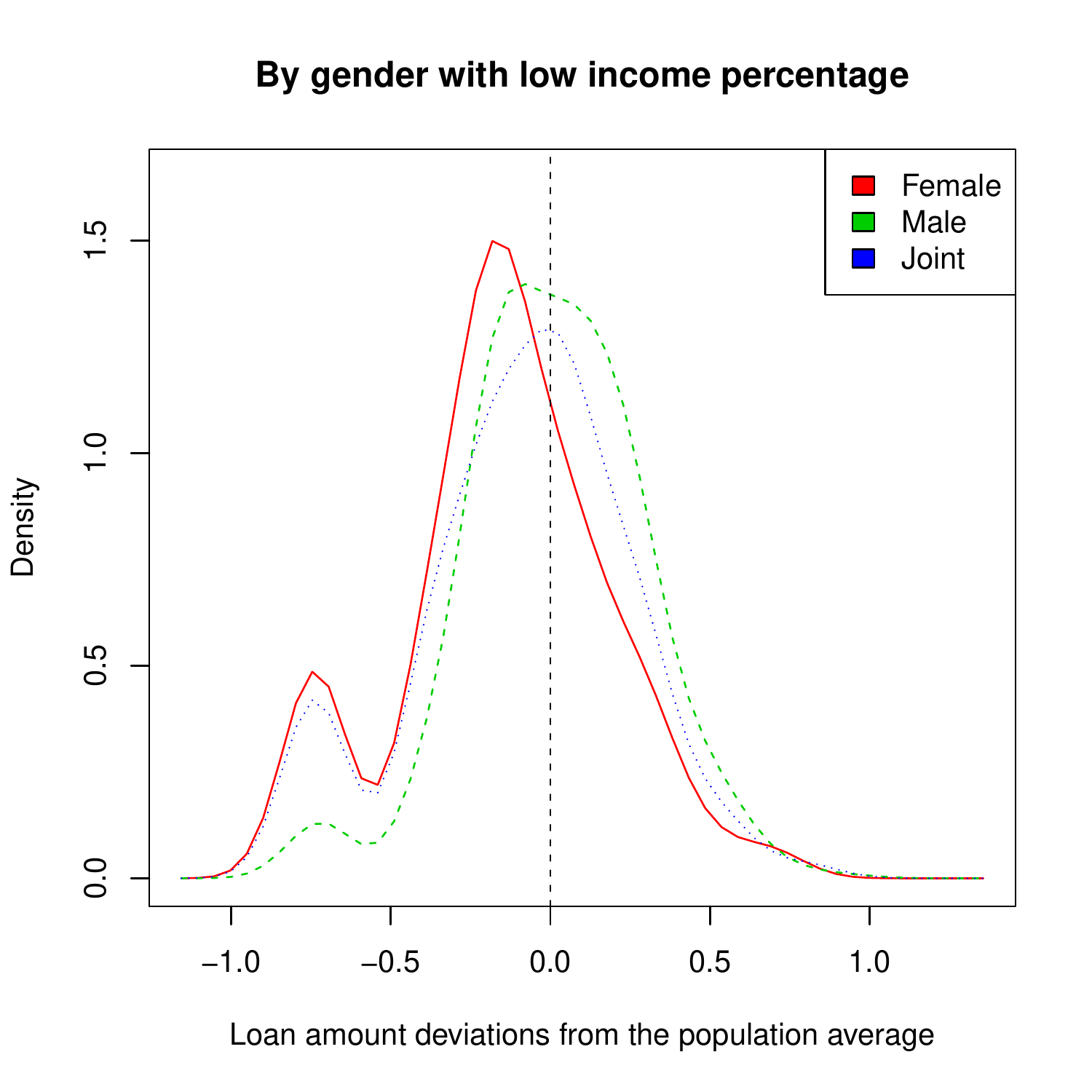} 
\caption{VT loan amount deviations from the population mean based on full data and sub-populations with low/high minority and low/high income percentages.}
\label{fig:HDMA_DU_VT}
\end{figure}

Figures \ref{fig:HDMA_DU_ME} and \ref{fig:HDMA_DU_VT} show the kernel
densities of the deviation of the log-transformed loan amount from the
population mean. In particular, we plot such deviations for the entire
dataset, but also for sub-populations with low/high minority and income
percentage. Low/high minority percentage means that the mortgage applicant
is in an area of lower/higher minority than the population median. Moreover,
low/high income percentage means that the mortgage applicant household
income is lower/higher than the household income in her/his MSA. ME data
from Figures \ref{fig:HDMA_DU_ME} do not show any evidence of DU with
respect to the loan amount, which explains the results in Table~\ref%
{denoiseresults--2} where SP-SVM and EEL-SVM did not improve the non-robust
counterpart. VT data show a very different scenario in Figure~\ref%
{fig:HDMA_DU_VT}, where the loan amount deviations have a bimodal
distribution. In addition, within the low minority sub-population, female
applicants exhibit significantly lower loan amounts than all other
applicants; the same pattern is observed in the low income sub-population.
Therefore, the DU in the VT data is evident, and confirms the findings in
Table~\ref{tab:KS-finance}, but also those in Table~\ref{denoiseresults--2}
where SP-SVM and EEL-SVM did improve the non-robust C-SVM.

We have concluded the first part of our qualitative analysis, where we have
explained the DU, and we now evaluate how much compliant the automatized
mortgage lending process would be for the VT data; we do not report the ME
results due to lack of DU. Fairness compliance requires the lending decision
-- especially the unfavorable decisions ($Y=-1$) -- to be independent of the
applicant's gender at birth information, i.e. 
\begin{equation}
\Pr(Y=-1)=\Pr(Y=-1|S=l)\quad \text{for all}\quad l\in \mathcal{S}%
:=\{Female,Male,Joint\}.  \label{indep_Y}
\end{equation}

\begin{table}[H]
\begin{center}
\begin{tabular}{lrrr|r}
\hline
& C-SVM & SP-SVM & EEL-SVM & True \\ \hline
$\Pr(Y = -1)$ & 2.3148\% & 1.0802\% & \textbf{5.5556\%} & 7.8704\% \\ \hline
$\Pr(Y = -1|\text{Female})$ & 3.9683\% & 1.5873\% & \textbf{6.3492\%} & 
9.5238\% \\ 
$\Pr(Y = -1|\text{Male})$ & 7.8125\% & 3.9063\% & \textbf{10.1563\%} & 
16.4062\% \\ 
$\Pr(Y = -1|\text{Joint})$ & 0.0000\% & 0.0000\% & \textbf{3.8071\%} & 
4.5685\% \\ \hline
$\Pr(Y = -1|\text{Low\ Income})$ & 2.9316\% & 0.9772\% & \textbf{5.5375\%} & 
8.0495\% \\ 
$\Pr(Y = -1|\text{High\ Income})$ & 1.7595\% & 1.1730\% & \textbf{5.5718\%} & 
7.6923\% \\ \hline
$\Pr(Y = -1|\text{Low\ Minority})$ & 1.8576\% & 0.9290\% & \textbf{5.5728\%}
& 9.7720\% \\ 
$\Pr(Y = -1|\text{High\ Minority})$ & 2.7692\% & 1.2308\% & \textbf{5.5385\%}
& 6.1584\% \\ \hline
$CDD$ & -25.5944\% & -25.2601\% & \textbf{-8.3263\%} & -11.3792\% \\ \hline
\end{tabular}
\end{center}
\caption{Probability of a denied loan for the three classification methods.
The closest value (per row) to the `true' probability is in bold.}
\label{tab:probabilities}
\end{table}

Table~\ref{tab:probabilities} tells us that the desirable lack of disparity
in \eqref{indep_Y} is achieved best by EEL-SVM. In addition, EEL-SVM
estimates the unfavorable decisions very similarly to the `true' decisions
and this could be seen by inspecting the probabilities that appear in bold.

One popular fairness metric is the \emph{Conditional Demographic Disparity
(CDD)}, which is discussed in \cite{Wachter_et_al_2021}, where the data are
assumed to be part of multiple strata. The CDD formulation for our data
(with three strata, i.e. Female, Male and Joint) is defined as%
\begin{equation*}
CDD=\sum_{l\in \mathcal{S}}\Pr (S=l)\times DD_{l},
\end{equation*}%
where $DD_{l}$ is the \emph{demographic disparity} within the $l^{th}$
stratum, i.e. 
\begin{equation*}
DD_{l}=\Pr (S=l\mid Y=-1)-\Pr (S=l\mid Y=1)\quad \text{for all}\quad l\in 
\mathcal{S}.
\end{equation*}%
CDD could capture and explain peculiar data behavior similar to Simpson's
paradox where the same trend is observed in each stratum, but the opposite
trend is observed in the whole dataset. Amazon SageMaker, a cloud
machine-learning platform developed by Amazon, has included CDD in their
practice to enhance model explainability and bias detection; for details,
see the Amazon SageMaker Developer Guide.

Table~\ref{tab:probabilities} shows that EEL-SVM has superior performance to
C-SVM and SP-SVM when looking at the overall CDD fairness performance. In
fact, EEL-SVM exhibits fairer post-training decisions than the pre-training
fairness measured on the `true' mortgage lending decisions observed in the
testing data. In summary, the unanimous conclusion is that EEL-SVM shows the
fairest and most robust mortgage lending automatized decision.

\section{Conclusions and Future Work\label{concl sec}}
This paper addresses the problem of binary classification under DU. Two robust SVM-like classification algorithms are developed, namely, SP-SVM and EEL-SVM, and there is sufficient empirical evidence to conclude that the new classifiers are very competitive when compared to other well-known SVM robust competitors.  One could argue that SP-SVM shows slightly better classification performance than EEL-SVM and various alternatives, but there is an irrefutable evidence about the computational time, with both SP-SVM and EEL-SVM exhibiting significant competition advantage over all other choices.

Future research would be required to extend our probabilistic arguments to multi-class classification problems, which is a non-trivial problem. We also plan to adapt our SP-SVM and EEL-SVM to imbalanced data, which is another research direction for future investigation.

\bibliographystyle{rss}

\appendix



\renewcommand*{\thesection}{A.\arabic{section}} \renewcommand*{\theequation}{%
A.\arabic{equation}}

\section{Appendix\label{app sec}}

\subsection{Proof of Theorem~\ref{Fisher class SVM th}}\label{proof_th1}

It is sufficient to show that \eqref{Fisher class SVM def} holds when $p>q$ and $p<q$, where
\[p:=\Pr\big(Y=1|\textbf{x}\big)\quad\mbox{and}\quad q:=\Pr\big(Y=-1|\textbf{x}\big).\]
The latter is equivalent to 
\begin{eqnarray}\label{Fisher class SVM th: eq1}
	\argmin_{z\in\mathbb{R}} \mathbf{E}_{\mathcal{Y}|\textbf{x}} L\Big(1\!-\!Y z\Big)
	= \argmin_{z\in\mathbb{R}} p L(1\!-\!z)+q L(1\!+\!z)
	=\left\{ \begin{array}{rll}
		1, & \text{if}\; & p>q, \\
		-1, & \text{if}\; & p<q.
	\end{array} \right.
\end{eqnarray}
Note that $L(1\pm\cdot)$ are compositions of the convex function $L$ with affine mappings, and therefore, the objective function of \eqref{Fisher class SVM th: eq1} is convex. Moreover, the left and right derivatives of $L$ exist as the loss function is convex. 

Assume first that $p>q$. The left and right derivatives at $1$ of the objective function in \eqref{Fisher class SVM th: eq1} are $-p L'(0^+)+q L'(2^-)$ and $-p L'(0^-)+q L'(2^+)$, respectively. Clearly,
\[-p L'(0^+)+q L'(2^-)=L'(0^+)(q-p)\le0\]
is true as $L$ is linear on $(0,2+\epsilon)$ for some $\epsilon>0$. Further,
\[\quad-p L'(0^-)+q L'(2^+)\ge 0\]
also holds due to the fact that $L'(0^-)\le 0\le L'(2^+)$, which is a consequence of the convexity of $L$ that attains its global minimum at $0$. Thus, the global minimum of \eqref{Fisher class SVM th: eq1} is attained at $1$ whenever $p>q$.

Assume now that $p<q$. Similarly, the left and right derivatives at $-1$ of the objective function in \eqref{Fisher class SVM th: eq1} are $-p L'(2^+)+q L'(0^-)$ and $-p L'(2^-)+q L'(0^+)$, respectively. Clearly, $-p L'(2^+)+q L'(0^-)\le0$ holds as $L'(0^-)\le 0\le L'(2^+)$ and $L$ is convex attaining its global minimum at $0$.  Further, $-p L'(2^-)+q L'(0^+)=L'(0^+)(q-p)\ge0$ is true as $L$ is linear on $(0,2+\epsilon)$ for some $\epsilon>0$. Thus, the global minimum of  \eqref{Fisher class SVM th: eq1} is attained at $-1$ whenever $p<q$. This completes the proof.

\subsection{Explicit Solution for (\ref{SVM non_sep simple})\label{sol_SP_SVM}}

Let $\phi_j\big(\mathbf{x}_i\big)$ be the $j^{th}$ element of $\phi\big(\mathbf{x}_i\big)$. Denote by $\phi_1\big(\mathbf{x}_i\big)$ and $\phi_2\big(\mathbf{x}_i\big)$ two vectors with their $j^{th}$ elements given by
$\phi_{1j}\big(\mathbf{x}_i\big)=\phi_j\big(\mathbf{x}_i\big)-a_{ik}I_{j=k}$ and $\phi_{2j}\big(\mathbf{x}_i\big)=\phi_j\big(\mathbf{x}_i\big)+a_{ik}I_{j=k}$ for all $1\le i\le N$ and $1\le j\le d$, where $I_{A}$ is the indicator of set $A$ that takes the values $1$ or $0$ if $A$ is true or false, respectively. Thus, \eqref{SVM non_sep simple} could be written as
\begin{eqnarray}\label{SVM non_sep simple: v2}
	\begin{array}{lllrl}
		&\underset{\mathbf{w}, b, \boldsymbol{\xi}}{\text{min}} & \frac{1}{2}\mathbf{w}^T \mathbf{w} + C \displaystyle\sum_{i=1}^N \xi_i &&\\
		& \;\;\text{s.t.}& y_i\Big(\mathbf{w}^T \phi\big(\mathbf{x}_i\big)+b\Big)\geq 1 - \xi_i, &\xi_i \geq 0, &1\le i\le N,\\
		& & y_i\big(\mathbf{w}^T \phi_k(\mathbf{x}_i)+b\big)\geq 1 - \xi_i,& k\in\{1,2\}, & 1\le i\le N.
	\end{array}
\end{eqnarray}
It should be noted that the above is a convex quadratic optimization problem that has only affine constraints, and thus, strong duality holds. 
The dual of \eqref{SVM non_sep simple: v2} is given by
\begin{eqnarray}\label{rob SVM: eq2}
	\begin{array}{llll}
		&\underset{\boldsymbol{\alpha},\boldsymbol{\beta},\boldsymbol{\gamma},\boldsymbol{\delta}\ge\textbf{0}}{\text{max}} & -\frac{1}{2}\big[\boldsymbol{\alpha}\;\boldsymbol{\beta}\;\boldsymbol{\gamma}\big]^T \;\textbf{T}\; \big[\boldsymbol{\alpha}\; \boldsymbol{\beta}\;\boldsymbol{\gamma}\big]+ \textbf{1}^T\boldsymbol{\alpha}+\textbf{1}^T\boldsymbol{\beta}+\textbf{1}^T\boldsymbol{\gamma} &\\
		& \;\;\;\;\;\text{s.t.}                               & 
		\boldsymbol{\alpha}+\boldsymbol{\beta}+\boldsymbol{\gamma}+\boldsymbol{\delta}= C\textbf{1},&\\
		&                                           & \textbf{y}^T\boldsymbol{\alpha}+\textbf{y}^T\boldsymbol{\beta}+\textbf{y}^T\boldsymbol{\gamma}=0,  &
	\end{array}
\end{eqnarray}
where the block matrix $\textbf{T}$ is given by
\begin{eqnarray*}
	\textbf{T}=\left[
	\begin{array}
		{c|c|c}
		\textbf{T}^{\phi,\phi} & \textbf{T}^{\phi,\phi_1} & \textbf{T}^{\phi,\phi_2}\\
		\hline
		\textbf{T}^{\phi_1,\phi} & \textbf{T}^{\phi_1,\phi_1} & \textbf{T}^{\phi_1,\phi_2}\\
		\hline
		\textbf{T}^{\phi_2,\phi} & \textbf{T}^{\phi_2,\phi_1} & \textbf{T}^{\phi_2,\phi_2}
	\end{array}\right]
\end{eqnarray*}
with $\textbf{T}^{\varphi_1,\varphi_2}$ being an $N\times N$ matrix with the $(i,j)^{th}$ entry given by $y_i\varphi_1^T(\mathbf{x}_i)\varphi_2(\mathbf{x}_i)y_j$ for all $\varphi_1,\varphi_2\in\{\phi,\phi_1,\phi_2\}$ and $1\le i,j\le N$. Clearly, \eqref{rob SVM: eq2} is equivalent to solving 
\begin{eqnarray}\label{rob SVM: eq3}
	\hspace{-0.5cm}\begin{array}{llll}
		&\underset{\boldsymbol{\alpha},\boldsymbol{\beta},\boldsymbol{\gamma}\ge\textbf{0}}{\text{min}} & \frac{1}{2}\big[\boldsymbol{\alpha}\;\boldsymbol{\beta}\;\boldsymbol{\gamma}\big]^T \;\textbf{T}\; \big[\boldsymbol{\alpha}\;\boldsymbol{\beta}\;\boldsymbol{\gamma}\big]-\textbf{1}^T\boldsymbol{\alpha}-\textbf{1}^T\boldsymbol{\beta}-\textbf{1}^T\boldsymbol{\gamma} &\\
		& \;\;\;\;\text{s.t.}                               & 
		\boldsymbol{\alpha}+\boldsymbol{\beta}+ \boldsymbol{\gamma}\le C\textbf{1},&\\
		&                                           & \textbf{y}^T\boldsymbol{\alpha}+\textbf{y}^T\boldsymbol{\beta}+\textbf{y}^T\boldsymbol{\gamma}=0.  &
	\end{array}
\end{eqnarray}

Let $(\boldsymbol{\alpha}^*,\boldsymbol{\beta}^*,\boldsymbol{\gamma}^*)$ be an optimal solution of \eqref{rob SVM: eq3}, which now helps with finding an optimal solution of \eqref{SVM non_sep simple}, which, in turn, gives us the classification rule identified by $\textbf{w}^*$ and $b^*$.  Clearly,
\begin{eqnarray*}
	\textbf{w}^*:=\sum_{i=1}^N \big(\alpha_i^* y_i\phi(\mathbf{x}_i)+\beta_i^* y_i\phi_1(\mathbf{x}_i)+\gamma_i^* y_i\phi_2(\mathbf{x}_i)\big).
\end{eqnarray*}
The choice of $b^*$ is possible by considering the complementary slackness conditions of \eqref{SVM non_sep simple: v2}.  A sensible estimate of $b^*$ is 
$\widehat{\hspace{1pt}b^*}: =\widehat{\hspace{1pt}b^+_l}/|\mathcal{S}_l|$,
where $|\mathcal{S}_l|$ represents the cardinality of $\mathcal{S}_l$ which is the set with the largest cardinality among
\begin{eqnarray*}
	&&\mathcal{S}_k:=\big\{1\le i\le N:\;\theta_{ik}^*(C-\alpha_i-\beta_i-\gamma_i)>0\big\},
\end{eqnarray*}
where $\theta_{i0}^*=\alpha_i^*$, $\theta_{i1}^*=\beta_i^*$ and $\theta_{i2}^*=\gamma_i^*$ for all $1\le i\le N$,  and 
\begin{eqnarray*}
	\widehat{\hspace{1pt}b^+_l}:=\!\sum_{j\in\mathcal{S}_l}y_j-\!\sum_{j\in\mathcal{S}_l}\sum_{i=1}^N \big(\alpha^*_i y_i \phi^T(\mathbf{x}_i)\phi(\mathbf{x}_j)
	+\beta^*_i y_i \phi_1^T(\mathbf{x}_i)\phi (\mathbf{x}_j)
	+\gamma^*_i y_i \phi_2^T(\mathbf{x}_i)\phi (\mathbf{x}_j)\big).
\end{eqnarray*}
\ignore{
	\begin{eqnarray*}
		&&b^*_:=\sum_{j\in\mathcal{S}}y_j-\sum_{j\in\mathcal{S}} \sum_{i=1}^N \alpha^*_i y_i \phi^T\big(\mathbf{x}_i\big)\phi\big(\mathbf{x}_j)\\
		&&\hspace{1cm}-\sum_{j\in\mathcal{S}} \sum_{i=1}^N \beta^*_i y_i \phi_1^T \big(\mathbf{x}_i\big)\phi \big(\mathbf{x}_j)\\
		&&\hspace{1cm}-\sum_{j\in\mathcal{S}} \sum_{i=1}^N \gamma^*_i y_i \phi_2^T \big(\mathbf{x}_i\big)\phi \big(\mathbf{x}_j\big)\\
		&&b^*_2=\sum_{j\in\mathcal{S}_1}y_j-\sum_{j\in\mathcal{S}_1} \sum_{i=1}^N \alpha^*_i y_i \phi^T\big(\mathbf{x}_i\big)\phi_1\big(\mathbf{x}_j)\\
		&&\hspace{1cm} -\sum_{j\in\mathcal{S}_1} \sum_{i=1}^N \beta^*_i y_i \phi_1^T \big(\mathbf{x}_i\big)\phi_1 \big(\mathbf{x}_j )\\
		&&\hspace{1cm}-\sum_{j\in\mathcal{S}_1} \sum_{i=1}^N \gamma^*_i y_i \phi_2^T \big(\mathbf{x}_i\big)\phi_1 \big(\mathbf{x}_j\big)\\
		&&b^*_3=\sum_{j\in\mathcal{S}_2}y_j-\sum_{j\in\mathcal{S}_2} \sum_{i=1}^N \alpha^*_i y_i \phi^T\big(\mathbf{x}_i\big)\phi_2\big(\mathbf{x}_j)\\
		&&\hspace{1cm} -\sum_{j\in\mathcal{S}_2} \sum_{i=1}^N \beta^*_i y_i \phi_1^T \big(\mathbf{x}_i\big)\phi_2 \big(\mathbf{x}_j )\\
		&&\hspace{1cm} -\sum_{j\in\mathcal{S}_2} \sum_{i=1}^N \gamma^*_i y_i \phi_2^T \big(\mathbf{x}_i\big)\phi_2 \big(\mathbf{x}_j\big)
\end{eqnarray*}}

\subsection{Explicit Solution for (\ref{EEL-SVM def})}\label{sol_EEL_SVM}

The derivations in this section are quite similar to those in Appendix \ref{sol_SP_SVM}, and thus, we provide only the main steps.  Note that the convex quadratic instance \eqref{EEL-SVM def} has only affine constraints, and therefore, the strong duality holds.  
\ignore{Now, the Lagrangian of \eqref{EEL-SVM def} is given by
	\begin{eqnarray}\label{rob EEL-SVM: eq1}
		\lefteqn{H\big(\mathbf{w}, b, \boldsymbol{\xi},z,\boldsymbol{\alpha},\boldsymbol{\beta},\boldsymbol{\gamma}\big)}\\
		&&:=\frac{1}{2}\mathbf{w}^T \mathbf{w} + D z+ \displaystyle\sum_{i=1}^N (D_1-\beta_i-\gamma_i)\xi_i-z\sum_{i=1}^N \beta_i\nonumber\\
		&&\hspace{0.5cm}-\sum_{i=1}^N \alpha_i\Big(y_i\big(\mathbf{w}^T \phi(\mathbf{x}_i)+b\big)+z-1 + \xi_i\Big),\nonumber
	\end{eqnarray}
	where $D_1:=\frac{D}{N(1-\alpha)}$ and $\boldsymbol{\alpha},\boldsymbol{\beta},\boldsymbol{\gamma}\ge\textbf{0}$. 
	Due to the strong duality property, we only need to solve the dual of \eqref{EEL-SVM def}, and we thus set
	\begin{eqnarray*}
		&&\hspace{-0.7cm}\textbf{0}\!=\!\frac{\partial H}{\partial \textbf{w}}\!=\!\textbf{w}\!-\!\sum_{i=1}^N \alpha_i y_i\phi(\mathbf{x}_i),\;\; 0\!=\!\frac{\partial H}{\partial b}=-\textbf{y}^T\boldsymbol{\alpha},\\
		&&\hspace{-0.7cm}\textbf{0}\!=\!\frac{\partial H}{\partial \boldsymbol{\xi}}\!=\!D_1\textbf{1}\!-\!\boldsymbol{\alpha}-\boldsymbol{\beta}\!-\!\boldsymbol{\gamma},\quad\; 0\!=\!\frac{\partial H}{\partial z}\!=\!D\!-\!\textbf{1}^T\boldsymbol{\alpha}\!-\!\textbf{1}^T\boldsymbol{\beta}.
\end{eqnarray*}}
One may show that the dual of \eqref{EEL-SVM def} is equivalent to solving 
\begin{eqnarray}\label{rob EEL-SVM: eq2}
	\hspace{-0.5cm}\begin{array}{llll}
		&\underset{\boldsymbol{\alpha},\boldsymbol{\beta},\boldsymbol{\gamma}\ge\textbf{0}}{\text{min}} & \frac{1}{2}\boldsymbol{\alpha}\;\textbf{T}^{\phi,\phi}\;\boldsymbol{\alpha}-\textbf{1}^T\boldsymbol{\alpha} &\\
		& \;\;\text{s.t.}  & 
		\boldsymbol{\alpha}+\boldsymbol{\beta}+\boldsymbol{\gamma}= D_1\textbf{1},\\
		&&\textbf{y}^T\boldsymbol{\alpha}=0,\\
		&&\textbf{1}^T \boldsymbol{\alpha}+\textbf{1}^T \boldsymbol{\beta}=D,
	\end{array}
\end{eqnarray}
where $\textbf{T}^{\phi,\phi}$ is as defined in Appendix \ref{sol_SP_SVM} and $D_1:=D/N(1-\alpha)$. Once again, \eqref{EEL-SVM def} and \eqref{rob EEL-SVM: eq2} are equivalent due to strong duality arguments.

Let $(\boldsymbol{\alpha}^*,\boldsymbol{\beta}^*,\boldsymbol{\gamma}^*)$ be an optimal solution of \eqref{rob EEL-SVM: eq2}. Then, \eqref{EEL-SVM def} is solved with
\begin{eqnarray*}
	\textbf{w}^*=\sum_{i=1}^N \alpha_i^* y_i\phi(\mathbf{x}_i).
\end{eqnarray*}
Finally, the bias term $b^*$  could be estimated as follows
\begin{equation*}
	\big(\widehat{\hspace{1pt}b^{*}},\widehat{\hspace{1pt}z^{*}}\big):=\left\{
	\begin{array}{cc}
		\big(\widehat{\hspace{1pt}b^{*1}},0\big)  & \text{if} \quad |\mathcal{S}_4|\le |\mathcal{S}_3| ,\\
		\big(\widehat{\hspace{1pt}b^{*2}},0\big)  & \text{if} \quad |\mathcal{S}_4|  > |\mathcal{S}_3|,
	\end{array}\right.
\end{equation*}
where
\begin{eqnarray*}
	&&\widehat{\hspace{1pt}b^{*1}}\!=\!\frac{\widehat{\hspace{1pt}b^+_3}}{|\mathcal{S}_3|}, \;\;\widehat{\hspace{1pt}b^+_3}\!=\!\sum_{j\in\mathcal{S}_3}y_j\!-\!\sum_{j\in\mathcal{S}_3} \sum_{i=1}^N \alpha^*_i y_i \phi^T(\mathbf{x}_i)\phi(\mathbf{x}_j),\\
	&&\widehat{\hspace{1pt}b^{*2}}\!=\!\frac{\widehat{\hspace{1pt}b^+_4}}{|\mathcal{S}_4|}, \;\;\widehat{\hspace{1pt}b^+_4}\!=\!\sum_{j\in\mathcal{S}_4}y_j\!-\!\sum_{j\in\mathcal{S}_4} \sum_{i=1}^N \alpha^*_i y_i \phi^T(\mathbf{x}_i)\phi(\mathbf{x}_j),
\end{eqnarray*}
and 
\begin{eqnarray*}
	\mathcal{S}_3:=\big\{1\le i\le N:\;\alpha_i^*\beta_i^*\gamma_i^*>0\big\}\quad\text{and}\quad\mathcal{S}_4:=\big\{1\le i\le N:\;\alpha_i^*\beta_i^*>0,\gamma_i^*=0\big\}.
\end{eqnarray*}

\end{document}